\pdfoutput=1
\def\APPONECOL{}
\documentclass[sigconf,nonacm]{acmart}

\usepackage{booktabs}
\usepackage{array}
\usepackage{float}
\newsavebox{\fitbox}
\newenvironment{fitw}[1]{\def\fitwidth{#1}\begin{lrbox}{\fitbox}}{\end{lrbox}%
  \typeout{FITW table \thetable: \the\wd\fitbox\space vs \the\dimexpr\fitwidth\relax}%
  \ifdim\wd\fitbox>\fitwidth\resizebox{\fitwidth}{!}{\usebox{\fitbox}}\else\usebox{\fitbox}\fi}
\usepackage{colortbl}
\definecolor{lightyellow}{RGB}{255,249,219}
\usepackage{multirow}
\usepackage{graphicx}
\usepackage{amsmath}
\usepackage{pifont}
\newcommand{\yes}{\ding{51}}
\newcommand{\no}{\ding{55}}
\providecommand{\xmark}{\ding{55}}

\usepackage{enumitem}

\setlist[itemize]{leftmargin=1.1em, labelsep=0.45em, itemsep=2pt, topsep=3pt,
                  parsep=0pt}

\definecolor{newred}{RGB}{0,0,0}
\newcommand{\new}[1]{{\color{newred}#1}}
\newcommand{\revcolor}{\color{newred}}
\definecolor{rtwored}{RGB}{0,0,0}
\newcommand{\rtwo}[1]{{\color{rtwored}#1}}
\AtBeginDocument{\pdfstringdefDisableCommands{\def\new#1{#1}\def\rtwo#1{#1}\def\revcolor{}}}
\definecolor{bestred}{RGB}{139,0,0}
\newcommand{\best}[1]{\textbf{\color{bestred}#1}}

\newcommand{\ArticlesPerDay}{99}
\newcommand{\NAltCkpt}{44}
\newcommand{\NDomain}{18}
\newcommand{\NValChrono}{6}

\newcommand{\NValDated}{2}
\newcommand{\NValInstr}{4}

\newcommand{\SelSetting}{rank 4 with 1200 steps for ChronoGPT, rank 1 with 600 steps for ChronoGPT-Instruct, and rank 16 with 300 steps for DatedGPT, all at a learning rate of $10^{-4}$}

\newcommand{\ShareDupText}{4.9\%}
\newcommand{\ShareMidnight}{99.6\%}
\newcommand{\ShareMultiArticle}{12\%}
\newcommand{\TickersPerDay}{86}

\usepackage{etoolbox}
\AtBeginEnvironment{table}{\setlength{\abovecaptionskip}{9pt}\setlength{\belowcaptionskip}{9pt}}
\AtBeginEnvironment{table*}{\setlength{\abovecaptionskip}{9pt}\setlength{\belowcaptionskip}{9pt}}
\AtBeginEnvironment{figure}{\setlength{\abovecaptionskip}{9pt}\setlength{\belowcaptionskip}{9pt}}
\AtBeginEnvironment{figure*}{\setlength{\abovecaptionskip}{9pt}\setlength{\belowcaptionskip}{9pt}}

\makeatletter
\newcommand{\figcaption}{\def\@captype{figure}\caption}
\newcommand{\tabcaption}{\def\@captype{table}\caption}
\makeatother

\renewcommand\footnotetextcopyrightpermission[1]{}

\begin{document}

\setcounter{dbltopnumber}{3}
\setcounter{topnumber}{3}
\setcounter{totalnumber}{6}
\renewcommand{\dbltopfraction}{0.92}
\renewcommand{\topfraction}{0.92}
\renewcommand{\textfraction}{0.06}
\renewcommand{\dblfloatpagefraction}{0.7}

\title{PALM: Point-in-Time Adaptation for Financial Language Models}

\author{Seunghan Lee}
\affiliation{%
  \institution{LG AI Research}
  \city{Seoul}
  \country{Republic of Korea}}
\author{Jun Seo}
\affiliation{%
  \institution{LG AI Research}
  \city{Seoul}
  \country{Republic of Korea}}
\author{Jaehoon Lee}
\affiliation{%
  \institution{LG AI Research}
  \city{Seoul}
  \country{Republic of Korea}}
\author{Junhyeok Kang}
\affiliation{%
  \institution{LG AI Research}
  \city{Seoul}
  \country{Republic of Korea}}
\author{Sangjun Han}
\affiliation{%
  \institution{LG AI Research}
  \city{Seoul}
  \country{Republic of Korea}}
\author{Sungdong Yoo}
\affiliation{%
  \institution{LG AI Research}
  \city{Seoul}
  \country{Republic of Korea}}
\author{Minjae Kim}
\affiliation{%
  \institution{LG AI Research}
  \city{Seoul}
  \country{Republic of Korea}}
\author{Tae Yoon Lim}
\affiliation{%
  \institution{LG AI Research}
  \city{Seoul}
  \country{Republic of Korea}}
\author{Dongwan Kang}
\affiliation{%
  \institution{LG AI Research}
  \city{Seoul}
  \country{Republic of Korea}}
\author{Hwanil Choi}
\affiliation{%
  \institution{LG AI Research}
  \city{Seoul}
  \country{Republic of Korea}}
\author{Soonyoung Lee}
\affiliation{%
  \institution{LG AI Research}
  \city{Seoul}
  \country{Republic of Korea}}
\author{Wonbin Ahn}
\authornote{Corresponding author.}
\affiliation{%
  \institution{LG AI Research}
  \city{Seoul}
  \country{Republic of Korea}}

\renewcommand{\shortauthors}{Lee et al.}

\begin{abstract}
Language models used in financial backtests suffer from \textit{look-ahead bias},
as a model trained on text
published after the study period
has already observed the outcomes it is asked to predict.
To handle this issue, \textit{point-in-time (PIT) language models} are pretrained on chronologically filtered corpora
and released as one checkpoint per calendar year,
each with a documented cutoff.
However, each additional year costs \new{another} pretraining run,
and whether that run is necessary has never been tested.
In this paper, we show that \new{\textit{a newer checkpoint is not \rtwo{detectably} better on the decision it informs}, as its information coefficient on a fixed evaluation window shows no decay with age that we can detect.} Motivated by this observation, we propose PALM (\textbf{P}oint-in-time \textbf{A}daptation for financial \textbf{L}anguage \textbf{M}odels), a simple yet effective alternative to annual pretraining that fits a low-rank adapter on \new{financial text the checkpoint is allowed to read,} without modifying any pretrained weight. \new{We further find that the gain comes from the financial domain rather than from recency, which leaves the point-in-time constraint without a detectable cost.} We validate PALM on a decade of financial news and on \new{three} families of PIT models, whose cutoffs span two decades. \new{The pooled gain keeps its sign across experimental settings and comes at a small fraction of the cost of pretraining.} Code is available at: \url{https://github.com/seunghan96/palm}.

\end{abstract}




\maketitle

{\let\thefootnote\relax\footnotetext{Accepted at the ICAIF '26 Workshop on LLMs and Generative AI for Finance (AI4F).}}

\section{Introduction}

Large language models (LLMs) have become a standard tool for extracting information from unstructured text, and finance is among the domains that have adopted them fastest \citep{wu2023bloomberggpt,li2023finllmsurvey}. Financial text arrives continuously as news articles, earnings calls, and regulatory filings, and prior work \citep{ke2019predicting,dong2024fnspid,wu2025press} shows that this text carries information about future returns. A common setup is to score each article with an LLM, rank assets on that score to form a cross-sectional portfolio, and evaluate the portfolio on historical data \citep{iacovides2024finllama,vamvourellis2025reasoning}.

\begin{figure}[t]
\centering
\includegraphics[width=0.958\columnwidth]{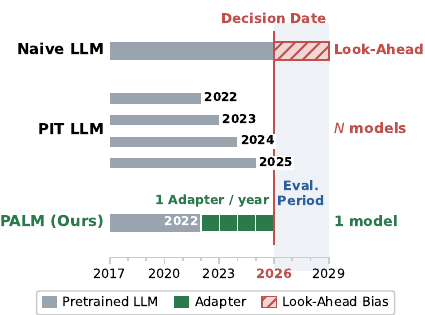}
\Description{A timeline from 2017 to 2029 with three rows. The naive model reads text past the decision date at 2026 and its overrun is hatched as look-ahead bias. The point-in-time suite is four bars, one per calendar year, each stopping before the decision date. Our row is one bar of the same pretrained text followed by four small adapter blocks, one per year.}
\caption{Naive LLM vs. PIT LLM vs. PALM (Ours). A naive LLM has read the evaluation period and is ineligible. A PIT suite avoids this with one pretraining run per year, whereas PALM needs one model and one small adapter \new{fitted on financial text that the model is allowed to read}.}
\label{fig:concept}
\end{figure}
However, a backtest is trustworthy only if the model has not already read the period it is tested on. As shown in Figure~\ref{fig:concept}, a model trained through 2026 has observed how the market responded to news published in 2022, and its score for that news recalls the outcome rather than forecasting it. This is known as \textit{look-ahead bias}, and it inflates backtested performance \citep{glasserman2023lookahead,lopez2026can,li2026oracle,kong2026evaluating}. The bias grows with every release, since a newer model has read more of whatever period a study tests, and its size cannot be measured, as a released model rarely documents either its corpus or its cutoff \citep{cheng2024dated}.

\begin{table}[t]
{\revcolor
\caption{\new{\textbf{Effectiveness \& Efficiency of PALM.} On DatedGPT, PALM reaches an IC comparable to full finetuning with a higher return and Sharpe ratio, showing that \textit{it avoids look-ahead bias far more cheaply}.}}
\label{tab:compute}
\small
\setlength{\tabcolsep}{4pt}
\begin{tabular}{ll|ccc}
\toprule
 & & w/o FT & Full FT & \cellcolor{lightyellow}\textbf{PALM} \\
\midrule
\multirow{5}{*}{Cost} & Trainable params & 0 & 1.35B (100\%) & \cellcolor{lightyellow}\textbf{6.05M (0.45\%)} \\
 & Tokens read & 0 & $\le$0.31M & \cellcolor{lightyellow}\textbf{$\le$0.15M} \\
 & Training time & \textemdash & 76\,s & \cellcolor{lightyellow}\textbf{16\,s} \\
 & Peak GPU memory & \textemdash & 20.9\,GB & \cellcolor{lightyellow}\textbf{12.0\,GB} \\
 & Storage per date & \textemdash & 2.7\,GB & \cellcolor{lightyellow}\textbf{0.02\,GB} \\
\midrule
\multirow{3}{*}{Effect} & IC ($\times10^{-3}$) & 0.17 & \textbf{0.81} & \cellcolor{lightyellow}0.80 \\
 & Return (\%) & $-$0.68 & 0.05 & \cellcolor{lightyellow}\textbf{0.46} \\
 & Sharpe & $-$0.15 & 0.01 & \cellcolor{lightyellow}\textbf{0.10} \\
\bottomrule
\end{tabular}
}
\end{table}

To eliminate this bias by construction, recent work has introduced point-in-time (PIT) language models \citep{sarkar2024lookahead}. Each model is pretrained only on text published before its own cutoff, and one such checkpoint, or \textit{vintage}, is released per calendar year with that cutoff documented \citep{he2025chrono,he2025instruct,kelly2026pit,yan2026datedgpt}. A backtest run at date $t$ uses a vintage whose cutoff precedes $t$, so it has never seen the test period.

Although this removes look-ahead, the guarantee is expensive to maintain, since each additional year of coverage requires \new{another} pretraining run. \new{As Table~\ref{tab:compute} previews, a small adapter is a far lighter alternative.} Each such family of vintages, or \textit{suite}, now spans two decades of yearly pretraining runs. These runs are justified by a single premise, that a checkpoint decays with its \textit{staleness}, the gap between its cutoff and the date of the backtest. However, this premise has not been tested, since each vintage is evaluated on the period that follows its own cutoff, which moves the checkpoint and the period together. As a result, \textit{a vintage is never compared against the vintage that replaced it}.

\new{However, measuring that premise on the decision itself raises a further problem. Most financial news carries a date but no time of day, so a score compared with the close-to-close return of its own date can read the overnight reaction to news published after the close, which no trader could have acted on.}

\new{In this paper, we hold the evaluation window fixed, score every vintage on next-day open-to-close returns, which remain tradable under date-only timestamps, and find that \textit{a newer vintage is not \rtwo{detectably} better}: the slope of the information coefficient on staleness is positive in its point estimate, and its interval rules out any decay larger than a tenth of the gain of PALM per year.} To this end, we propose \textbf{PALM} (\textbf{P}oint-in-time \textbf{A}daptation for financial \textbf{L}anguage \textbf{M}odels), a simple yet effective alternative to pretraining a new checkpoint, which fits a low-rank adapter on \new{financial text the checkpoint is allowed to read}. All it needs is text and the language modeling objective, and it leaves every pretrained weight frozen. The adapter reads no text from the evaluation window, and an adapted checkpoint therefore remains eligible whenever the original checkpoint is. \new{Full finetuning on the same text avoids look-ahead bias as well, but PALM does so at a comparable IC (Section~\ref{sec:fullft}), and on DatedGPT with 0.45\% of the weights, a fifth of the training time and under a hundredth of the storage (Table~\ref{tab:compute} and Appendix~\ref{app:cost}).} The main contributions are \new{summarized as follows}:

\begin{itemize}
\item To the best of our knowledge, we are the first to evaluate the necessity of the pretraining run behind a PIT suite, and we find that staleness does not degrade downstream performance \new{detectably, under dependence-aware inference}.
\item We propose PALM, a simple yet effective plug-in method that avoids look-ahead bias by training a low-rank adapter on eligible text instead of pretraining a new checkpoint.
\item \new{We show that the adapter does not need text from after the cutoff, which explains why a newer vintage is no better and leaves the point-in-time constraint without a detectable cost.}
\item We conduct extensive experiments across various families of PIT models \new{with the adapter setting chosen on a separate validation window, a tradable target under date-only timestamps, dependence-aware inference, and matched alternatives}.
\end{itemize}

\setcounter{figure}{2}
\begin{figure*}[t]
\setlength{\abovecaptionskip}{2pt}\setlength{\belowcaptionskip}{2pt}
\centering
\includegraphics[width=\textwidth]{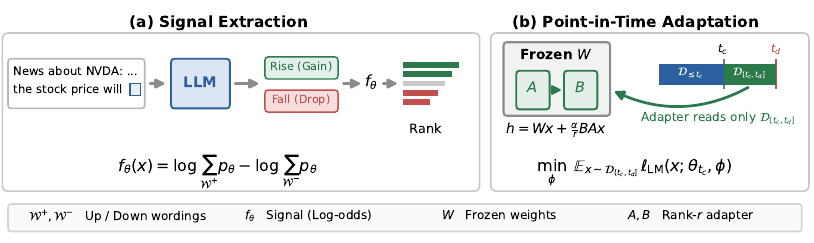}
\Description{Two boxed panels. The left panel takes a dated news article, ends the prompt where a direction word would follow, reads the log-odds between upward and downward continuations out of one forward pass, and ranks the assets of that date. The right panel freezes the pretrained weights and trains a rank-r adapter on the text published between the cutoff and the decision date, with an arrow showing that the adapter reads only that interval.}
\caption{Overall framework of PALM. (a) \textit{Signal Extraction}: A dated article is turned into a signal by contrasting upward against downward continuations in a single forward pass, and the signals are ranked within a date. The blue square marks the position whose next-token distribution is read. (b) \textit{Point-in-Time Adaptation}: Every pretrained weight is frozen and a low-rank adapter is fitted on the text published between the checkpoint cutoff and the decision date.}
\label{fig:method}
\end{figure*}
\setcounter{figure}{1}
\begin{figure}[t]
\centering
\includegraphics[width=\columnwidth]{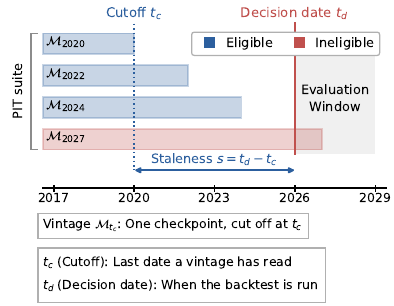}
\Description{A timeline from 2017 to 2029. Four horizontal bars, bracketed on the left as one point-in-time suite, each covering the text one vintage has read up to its own cutoff. Three bars stop before the decision date at 2026 and are marked eligible, and one runs past it into the shaded evaluation window and is marked ineligible. A dotted line marks the cutoff of the topmost vintage and a double arrow from it to the decision date is labelled staleness.}
\caption{Key concepts and notation. \new{Each vintage $\mathcal{M}_{t_c}$ of a PIT suite reads text only up to its cutoff $t_c$. It is eligible for the decision date $t_d$ when $t_c \le t_d$, and its staleness $s = t_d - t_c$ is the gap between the two. Every eligible vintage is scored on the same evaluation window after $t_d$.}}
\label{fig:setup}
\end{figure}
\setcounter{figure}{3}

\section{Related Works}

\begin{table}[t]
\caption{Positioning of PALM. Point-in-Time means that no text published after the decision date was read.}
\label{tab:novelty}
\small
\setlength{\tabcolsep}{4pt}
\begin{tabular}{l|ccc}
\toprule
Line of work & \begin{tabular}[c]{@{}c@{}}Point-\\in-Time\end{tabular} & Finance & Adaptation \\
\midrule
Financial LLMs \citep{ke2019predicting,lopez2026can,lee2026finstar} &      & \yes &      \\
PIT language models \citep{he2025chrono,kelly2026pit,yan2026datedgpt}                  & \yes & \yes &      \\
Continued pretraining \citep{lazaridou2021mind,dhingra2022time,luu2022time}            &      &      & \yes \\
Adapters \citep{houlsby2019adapter,hu2022lora}                   &      &      & \yes \\
\midrule
\rowcolor{lightyellow}
\textbf{PALM (Ours)}                                                                   & \yes & \yes & \yes \\
\bottomrule
\end{tabular}
\end{table}

\textbf{LLMs for financial text.} Textual sources such as news, earnings calls, and filings carry information about future returns, and extracting it has moved from lexicon counting to supervised models \citep{ke2019predicting} and now to LLMs \citep{lopez2026can,wu2023bloomberggpt,li2023finllmsurvey,lee2026finstar}. The standard evaluation scores each document, ranks assets cross-sectionally, and reports an information coefficient \citep{iacovides2024finllama,iacovides2025findpo,vamvourellis2025reasoning,wu2025press}. These studies are run on historical data with models whose cutoff is unknown, and PIT models were introduced to remove that uncertainty by construction.

\textbf{PIT language models.} Pretraining on date-filtered text has produced several public suites, including ChronoBERT and ChronoGPT \citep{he2025chrono}, PIT \citep{kelly2026pit}, and DatedGPT \citep{yan2026datedgpt}. They are evaluated on whether the resulting signal is comparable to that of an unconstrained model, but not on whether the vintage that replaced it is any better. A complementary line estimates the cutoff of a released model from dated text \citep{cheng2024dated}, which is necessary when it is not documented. We instead treat the documented cutoff as the experimental variable, varying it across vintages while the evaluation window is held fixed.

\textbf{Temporal degradation of LLMs.} Language models are known to degrade on text drawn from after their training window \citep{lazaridou2021mind,dhingra2022time,luu2022time}, an effect measured on perplexity and on knowledge probes \citep{loureiro2022timelms,zhu2025outdated}. That line of work evaluates the model rather than the decision, and proposes continued pretraining as the remedy\rtwo{, although part of the gain of such temporal adaptation has been traced to the domain rather than the date of the text \citep{rottger2021temporal,agarwal2022temporal}}. We instead measure it on the decision itself, scoring financial news and reporting the information coefficient a practitioner would act on.

\textbf{Parameter-efficient adaptation.} Low-rank adaptation \citep{hu2022lora,dettmers2023qlora} adds a small trainable module to a frozen model, mainly to reduce the cost of supervised finetuning \citep{houlsby2019adapter,li2021prefix,iacovides2024finllama}. A cheaper alternative is to leave the weights untouched and supply the missing text at inference time, as in retrieval augmentation \citep{lewis2020rag,izacard2023atlas,choi2025finder}, which under a point-in-time constraint needs an index restricted to eligible documents. We instead use an adapter to keep a checkpoint current rather than to finetune it cheaply, and the binding constraint is not the parameter budget but the publication date of the corpus. \new{The update is closest to domain-adaptive pretraining \citep{gururangan2020dont}, carried out with a low-rank adapter under a point-in-time rule.}

\textbf{Positioning of our work.} An ideal method of keeping a model current would have three properties at once: 1) point-in-time eligibility, 2) a focus on financial text, and 3) an update to the model itself. As summarized in Table~\ref{tab:novelty}, no line of work has all three. PALM combines them by fitting a parameter-efficient update on financial text the checkpoint was already allowed to read.

\section{Proposed Method: PALM}

\subsection{Problem Setup}

\textbf{Vintages.} Let $\mathcal{D} = \{(x_i, t_i)\}_{i=1}^{N}$ be a corpus of documents $x_i$ with publication dates $t_i$, and let $\mathcal{D}_{\le t_c} = \{(x_i, t_i) \in \mathcal{D} : t_i \le t_c\}$ denote its restriction to text published on or before a cutoff $t_c$. A PIT suite is a family $\{\mathcal{M}_{t_c}\}_{t_c \in \mathcal{C}}$ of checkpoints produced by a single training pipeline, where $\mathcal{M}_{t_c}$ carries parameters $\theta_{t_c}$ obtained by minimizing the language modeling loss on the truncated corpus,
\begin{equation}
\theta_{t_c} = \arg\min_{\theta} \; \mathbb{E}_{x \sim \mathcal{D}_{\le t_c}} \left[ \ell_{\mathrm{LM}}(x; \theta) \right] .
\label{eq:vintage}
\end{equation}
Producing one vintage therefore costs one pretraining run\new{, from scratch or from the last vintage}, and $|\mathcal{C}|$ grows by one every year.

\textbf{Eligibility and staleness.} A study conducted at a decision date $t_d$ may use $\mathcal{M}_{t_c}$ only when $t_c \le t_d$, since otherwise the model has read text published after the decision it informs. We call such a vintage \textit{eligible} for $t_d$, and define its \textit{staleness} as $s = t_d - t_c$, both of which Figure~\ref{fig:setup} draws. The prevailing convention is to select the eligible vintage of smallest staleness,
\begin{equation}
t_c^{\star} = \max \{ t_c \in \mathcal{C} : t_c \le t_d \} ,
\label{eq:select}
\end{equation}
\new{which is what makes each additional year worth the run in} Eq.~\eqref{eq:vintage}.

\textbf{Comparison design.} Testing whether a larger $t_c$ yields a better model requires holding everything else constant. However, existing evaluations move the window with $t_c$, and the two effects cannot be separated. We instead fix a decision date $t_d$, score every eligible vintage on that window, and regress performance on staleness $s$.

\subsection{Signal Extraction}

As shown in Figure~\ref{fig:method}, the method has two halves, one that reads a signal \new{from} a checkpoint and one that \new{adapts it under the PIT rule}.

\textbf{Scoring.} A checkpoint is judged by the signal it extracts from dated text. Let $x_{jt}$ be a document concerning asset $j$ and dated $t$, and let $\pi(x)$ denote a prompt that stops just before the word naming the direction of the move. The signal assigned to that document is read from a single forward pass as the log-odds between two sets of continuations,
\begin{equation}
f_\theta(x) = \log \!\!\sum_{w \in \mathcal{W}^{+}}\!\! p_\theta(w \mid \pi(x)) \; - \; \log \!\!\sum_{w \in \mathcal{W}^{-}}\!\! p_\theta(w \mid \pi(x)),
\label{eq:score}
\end{equation}
where $\mathcal{W}^{+}$ and $\mathcal{W}^{-}$ hold upward and downward wordings.

\textbf{Measuring the signal.} Signals are compared within a date rather than across dates, since the level of $f_\theta$ carries no meaning while its cross-sectional ordering does. Write $r_{j,t+1}$ for the realized \new{next-day open-to-close return of asset $j$ (Section~\ref{sec:align} explains this choice),} and $\rho$ for the Spearman correlation. The quality of the signal over a window $T$ is then the information coefficient
\begin{equation}
\mathrm{IC} = \frac{1}{|T|} \sum_{t \in T} \rho \big( \{ f_\theta(x_{jt}) \}_j , \; \{ r_{j,t+1} \}_j \big).
\label{eq:ic}
\end{equation}
Subtracting the mean within each date and taking the sign gives an equally weighted long-short portfolio, and we report its return and Sharpe ratio. The three measures are defined in detail in Section~\ref{sec:setup}.

\subsection{Point-in-Time Adaptation}

PALM \new{adapts an eligible vintage with the text available at} the decision date $t_d$ without changing a single pretrained weight. Let
\begin{equation}
\mathcal{D}_{(t_c,t_d]} = \{ (x_i, t_i) \in \mathcal{D} : t_c < t_i \le t_d \}
\label{eq:window}
\end{equation}
be the text published after the cutoff but before the decision date. The vintage has never read it, and a practitioner standing at $t_d$ already holds all of it.

\textbf{Adapter.} For each attention or feed-forward projection $W \in \mathbb{R}^{d_{\mathrm{out}} \times d_{\mathrm{in}}}$ of $\theta_{t_c}$ that it adapts (Appendix~\ref{app:impl}), PALM adds a low-rank update \citep{hu2022lora} beside it as
\begin{equation}
h = W x + \frac{\alpha}{r} B A x , \qquad A \in \mathbb{R}^{r \times d_{\mathrm{in}}}, \; B \in \mathbb{R}^{d_{\mathrm{out}} \times r} ,
\label{eq:lora}
\end{equation}
with rank $r \ll \min(d_{\mathrm{in}}, d_{\mathrm{out}})$ and a fixed scale $\alpha$. Both $A$ and $B$ are trained, and we initialize $B = 0$, which leaves the adapted model identical to the vintage before any update is applied.

\textbf{Objective.} Writing $\phi = \{ (A_\ell, B_\ell) \}_\ell$ for the adapter parameters across layers, PALM solves
\begin{equation}
\phi^{\star} = \arg\min_{\phi} \; \mathbb{E}_{x \sim \mathcal{D}_{(t_c,t_d]}} \left[ \ell_{\mathrm{LM}}(x; \theta_{t_c}, \phi) \right]
\label{eq:palm}
\end{equation}
with $\theta_{t_c}$ held frozen, and returns the adapted model $\mathcal{M}_{t_c \to t_d}$. Eq.~\eqref{eq:palm} is the same objective as Eq.~\eqref{eq:vintage}, differing only in the interval the corpus is drawn from and in which parameters are free. The adapter is far smaller, with $|\phi| \ll |\theta_{t_c}|$.

Two properties follow. First, $\mathcal{M}_{t_c \to t_d}$ is eligible for $t_d$ whenever $\mathcal{M}_{t_c}$ is, since neither Eq.~\eqref{eq:vintage} nor Eq.~\eqref{eq:palm} reads text published after $t_d$. Second, PALM is a plug-in method, as the update in Eq.~\eqref{eq:lora} sits inside existing linear layers and does not change the architecture.

\section{Experiments}

\subsection{Experimental Setup}
\label{sec:setup}

\begin{figure*}[!t]
\centering
\includegraphics[width=\textwidth]{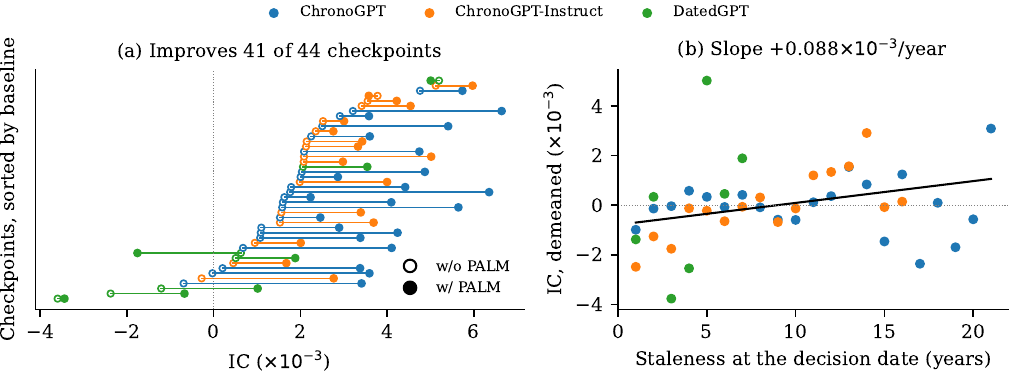}
\Description{Two panels. The left one connects each checkpoint to itself before and after the adapter, and almost every connector rises. The right one scatters the information coefficient against staleness for every eligible checkpoint of three suites, with a fitted line that does not slope downward.}
\caption{Effect of \new{PALM and staleness. Both panels read next-day open-to-close returns. (a) The same checkpoints before and after PALM, paired on the days both arms scored. (b) Information coefficient against staleness on the fixed evaluation window, which leaves the checkpoint as the only quantity varying between points. Points are demeaned within suite, as in Table~\ref{tab:stale}.}}
\label{fig:main}
\end{figure*}

\textbf{Datasets.} From FNSPID \citep{dong2024fnspid} we construct a panel of 252{,}000 article-day observations over 1{,}099 tickers between 2014 and 2023, sampled evenly across months. An article is credited to a trading day only if its timestamp precedes that day's close\new{, and is scored against the next-day open-to-close return, which is tradable whatever the hour of publication (Section~\ref{sec:align})}. The decision date is the end of November 2020, and every eligible vintage is scored on one window running from December 2020 through December 2023. \new{Table~\ref{tab:setup} summarizes the setup, Table~\ref{tab:data} details the panel, and Appendix~\ref{app:data} specifies the cross-section in full.}

\begin{table}[t]
\caption{Experimental setup: Data, Models, Evaluation.}
\label{tab:setup}
\small
\setlength{\tabcolsep}{4pt}
\begin{tabular}{l|l}
\toprule
Component & Setting \\
\midrule
\multicolumn{2}{l}{\textbf{[1] Data}} \\
\cmidrule(lr){1-2}
Corpus        & FNSPID financial news \citep{dong2024fnspid} \\
Universe      & 1{,}099 US tickers \\
Panel span    & 2014 to 2023 \\
Target        & \new{Next-day open-to-close return} \\
\midrule
\multicolumn{2}{l}{\textbf{[2] Models}} \\
\cmidrule(lr){1-2}
Checkpoints   & ChronoGPT \citep{he2025chrono}, ChronoGPT-Instruct \citep{he2025instruct} \\
              & DatedGPT \citep{yan2026datedgpt} \\
Cutoffs       & 1999 to 2024 \\
\midrule
\multicolumn{2}{l}{\textbf{[3] Evaluation}} \\
\cmidrule(lr){1-2}
Decision date & End of Nov 2020 \\
\new{Validation}    & \new{Dec 2017 to Nov 2020 (decision date end of Nov 2017)} \\
Evaluation    & Dec 2020 to Dec 2023 \\
Metric        & Information coefficient (IC), Return, Sharpe \\
\new{Inference}     & \new{Month-block $\times$ checkpoint bootstrap} \\
\bottomrule
\end{tabular}
\end{table}

\begin{table}[t]
\caption{\new{Details of FNSPID. Every eligible vintage is scored on the same articles of the evaluation window, showing that \textit{the comparison holds the data fixed across all checkpoints}.}}
\label{tab:data}
\small
\setlength{\tabcolsep}{4pt}
\resizebox{\columnwidth}{!}{%
\begin{tabular}{l|l}
\toprule
Property & Value \\
\midrule
\multicolumn{2}{l}{\textit{The panel}} \\
\cmidrule(lr){1-2}
Observations & 252{,}000 article-days over 2{,}514 trading days \\
Articles per day & 99 at the median, 100 on average \\
Article length & 417 characters at the median, 777 at the ninth decile \\
Publishers & 288 \\
\midrule
\multicolumn{2}{l}{\textit{The split at the decision date}} \\
\cmidrule(lr){1-2}
Readable by an adapter & 174{,}300 rows over 1{,}741 days, to 30 November 2020 \\
Evaluation window & 77{,}700 rows over 773 dates, 767 of them scored \\
\bottomrule
\end{tabular}}
\end{table}

\begin{table}[t]
\caption{Model checkpoints. Every suite releases one vintage per calendar year without gaps, and the counts differ only through where each suite starts and where it stops.}
\label{tab:ckpt}
\small
\setlength{\tabcolsep}{3.5pt}
\begin{tabular}{l|cc|cc|c}
\toprule
\multirow{2}{*}[-0.25\normalbaselineskip]{Suite} & \multicolumn{2}{c|}{Scored} & \multicolumn{2}{c|}{Eligible}
  & \multirow{2}{*}[-0.25\normalbaselineskip]{\begin{tabular}[c]{@{}c@{}}Excluded\\cutoffs\end{tabular}} \\
\cmidrule(lr){2-3}\cmidrule(lr){4-5}
 & $n$ & Cutoffs & $n$ & Cutoffs & \\
\midrule
ChronoGPT          & 26 & 1999--2024 & 21 & 1999--2019 & 2020 to latest \\
ChronoGPT-Instruct & 21 & 2004--2024 & 16 & 2004--2019 & 2020 to latest \\
DatedGPT           & 12 & 2013--2024 & 7  & 2013--2019 & 2020 to latest \\
\midrule
Total              & 59 & \textemdash & 44 & \textemdash & 15 \\
\bottomrule
\end{tabular}
\end{table}

\textbf{Backbones.} We use the public PIT suites that \new{release one vintage per calendar year}, and apply PALM to each without modifying the backbone. As shown in Table~\ref{tab:ckpt}, these span \new{two pipelines, one with an instruction-tuned variant,} and two decades of cutoffs. One further suite, PIT-4B \citep{kelly2026pit}, releases its vintages as \new{monthly} snapshots of one continuous run\new{, and we report it separately in Appendix~\ref{app:pit4b}}.

\textbf{Evaluation metrics.} We report three measures, each of which reads the same daily series in a different way.
\begin{itemize}
\item a) \textit{Information coefficient.} How well the score orders the cross-section on a given day, before any position is taken.
\item b) \textit{Annualized return.} What that ordering is worth once it is turned into an equally weighted long-short book.
\item c) \textit{Sharpe ratio.} What the same book earns per unit of risk\new{, the number a desk is held to and judged by}.
\end{itemize}
The first is Eq.~\eqref{eq:ic}, and for the other two we write the position taken on name $j$ at date $t$ as $w_{jt} = \operatorname{sign}(f_\theta(x_{jt}) - \bar f_t)$, with $\bar f_t$ the mean score on that date. The portfolio return is then $r_t = \sum_j w_{jt} r_{j,t+1} / \sum_j |w_{jt}|$, with mean $\bar r$ and standard deviation $\sigma_r$ over the window. We report $252\,\bar r$ as the annualized return and $\sqrt{252}\,\bar r / \sigma_r$ as the Sharpe ratio. Unless otherwise stated, the IC is in units of $10^{-3}$ and the annualized return in percentage points. Wins is the share of checkpoints whose measure the update raises, with the same checkpoint scored before and after. Both arms of a comparison are scored on the same span of the panel\new{, through the same code and on the same batches of articles, so the adapter is the only difference}.

\textbf{Implementation details.} Scores come from Eq.~\eqref{eq:score} throughout. Every backbone is loaded as released, which leaves the adapter as the only thing we vary\rtwo{, and Table~\ref{tab:impl} lists every fixed setting}. \new{PALM uses \SelSetting{}.} \new{Robustness checks that instead share one setting across backbones use rank 2 with 600 steps at $10^{-4}$, and are marked as such.} \new{The setting of each backbone is selected on a separate validation window over a grid fixed in advance (Appendix~\ref{app:select}), so the evaluation window takes no part in any choice. Every interval comes from a bootstrap that resamples months and, within each suite, checkpoints \citep{kunsch1989jackknife}, and a claim of no decline is stated as the largest fall its 90\% interval excludes, a one-sided 5\% test \citep{schuirmann1987tost}. Gains carry two-sided 95\% intervals and slopes 90\% intervals, and Wins count checkpoints scored on the same dates, not independent trials. \rtwo{The pooled IC gain is the primary measure, with the return and the Sharpe ratio as secondary measures.}}

\subsection{Effectiveness of PALM}
\label{sec:eff}

\begin{table}[t]
{\revcolor
\caption{\new{Effectiveness of PALM. Each backbone uses the setting selected for it on the validation window. The adapter raises the IC on every suite, showing that \textit{the gain does not depend on which model PALM is attached to}.}}
\label{tab:adapt}
\small
\setlength{\tabcolsep}{2.4pt}
\resizebox{\columnwidth}{!}{%
\begin{tabular}{l|l|ccc|cc}
\toprule
Backbone & & IC & Return & Sharpe & IC gain [95\% CI] & Wins \\
\midrule
ChronoGPT & Base & 1.67 & 2.01 & 0.42 & & \\
 & + PALM & \best{4.17} & \best{3.65} & \best{0.76} & \best{+2.50} [-0.77, +5.92] & \best{100\%} \\
\cmidrule(lr){1-7}
ChronoGPT-Instruct & Base & 2.21 & 2.95 & 0.61 & & \\
 & + PALM & \best{3.52} & \best{3.79} & \best{0.79} & \best{+1.31} [-0.26, +2.99] & \best{94\%} \\
\cmidrule(lr){1-7}
DatedGPT & Base & 0.17 & -0.68 & -0.15 & & \\
 & + PALM & \best{0.80} & \best{0.46} & \best{0.10} & \best{+0.62} [-1.55, +2.75] & \best{71\%} \\
\midrule
All backbones & Base & 1.63 & 1.93 & 0.40 & & \\
 & + PALM & \best{3.40} & \best{3.19} & \best{0.66} & \best{+1.77} [-0.15, +3.71] & \best{93\%} \\
\bottomrule
\end{tabular}}
}
\end{table}

\begin{figure}[t]
\centering
\includegraphics[width=\columnwidth]{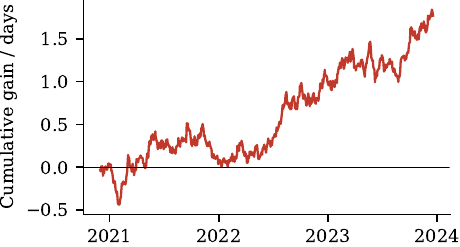}
\Description{The daily IC gain of PALM on next-day open-to-close returns, accumulated over the evaluation window and divided by its length, rising from zero at the start of the window to the mean gain at its end.}
\caption{\new{Timing of the gain. The daily IC gain on next-day open-to-close returns, accumulated over the evaluation window and divided by its length in units of $10^{-3}$, ends at the mean gain of Table~\ref{tab:adapt}, showing that \textit{the gain is not one episode}.}}
\label{fig:target}
\end{figure}

\begin{table*}[t]
{\revcolor
\caption{\new{Effect of staleness. Slopes are per year of staleness on next-day open-to-close returns, pooled with suite fixed effects and then fitted within each suite, with 90\% intervals from the month-block $\times$ checkpoint bootstrap. Every interval includes zero, and a decay is detected when the whole interval lies below zero, showing that \textit{no measure shows a decay we can detect}.}}
\label{tab:stale}
\small
\setlength{\tabcolsep}{4pt}
\begin{tabular}{l|c|ccc|ccc|ccc}
\toprule
\multirow{2}{*}[-0.5\normalbaselineskip]{Suite} & \multirow{2}{*}[-0.5\normalbaselineskip]{$n$} & \multicolumn{3}{c|}{IC} & \multicolumn{3}{c|}{Return} & \multicolumn{3}{c}{Sharpe} \\
\cmidrule(lr){3-5}\cmidrule(lr){6-8}\cmidrule(lr){9-11}
 & & Slope & 90\% CI & \begin{tabular}[c]{@{}c@{}}Decay\\detected\end{tabular} & Slope & 90\% CI & \begin{tabular}[c]{@{}c@{}}Decay\\detected\end{tabular} & Slope & 90\% CI & \begin{tabular}[c]{@{}c@{}}Decay\\detected\end{tabular} \\
\midrule
All suites (pooled) & 44 & +0.088 & [-0.140, +0.321] & \xmark & -0.116 & [-0.296, +0.071] & \xmark & -0.024 & [-0.063, +0.015] & \xmark \\
\midrule
ChronoGPT & 21 & +0.014 & [-0.259, +0.286] & \xmark & -0.187 & [-0.407, +0.029] & \xmark & -0.039 & [-0.083, +0.006] & \xmark \\
ChronoGPT-Instruct & 16 & +0.206 & [-0.077, +0.487] & \xmark & +0.012 & [-0.243, +0.271] & \xmark & +0.002 & [-0.052, +0.054] & \xmark \\
DatedGPT & 7 & +0.673 & [-0.444, +2.120] & \xmark & +0.292 & [-0.669, +1.390] & \xmark & +0.067 & [-0.144, +0.309] & \xmark \\
\bottomrule
\end{tabular}
}

\vspace{12pt}
\makeatletter\def\@captype{figure}\makeatother
\centering
\includegraphics[width=\textwidth]{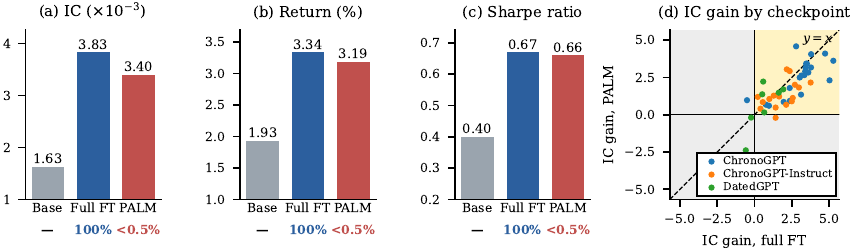}
\Description{Four panels. The first three are bars for the IC, the return and the Sharpe ratio pooled over the eligible checkpoints, each with the checkpoint as released, full finetuning and PALM; full finetuning and PALM reach similar heights, well above the release, and the line under the bars marks full finetuning as training 100 percent of the weights and PALM under 0.5 percent. The fourth scatters the IC gain of each checkpoint under PALM against that under full finetuning, with the diagonal; the points lie close to the diagonal.}
\caption{\new{PALM against full finetuning. (a)--(c) Pooled level of each measure, with the share of the weights each update trains under its bar. (d) IC gain of each checkpoint under the two updates, against the diagonal. PALM follows full finetuning on every measure and across checkpoints, showing that \textit{a low-rank adapter comes close to full finetuning at a fraction of its cost}.}}
\label{fig:fullft}
\end{table*}

To see whether an adapter can make effective use of \new{financial text} where the annual pretraining run cannot, we score each checkpoint with and without PALM and compare the two. As shown in Figure~\ref{fig:main}(a) and Table~\ref{tab:adapt}, PALM raises the IC on every backbone and on \new{most checkpoints. Under the bootstrap that resamples both months and checkpoints, the pooled gain on the IC is positive at the one-sided five percent level, and the gains in return and Sharpe ratio at the one-sided ten percent level, and Figure~\ref{fig:target} shows that it accrues mainly over 2022 and 2023 rather than in one stretch. Note that the settings are chosen without the evaluation window (Appendix~\ref{app:select}), so this gain is not tuned on the data it is reported on, and it survives removing the same-day and past-week return (Appendix~\ref{app:checks}).} \new{Note that ChronoGPT and DatedGPT share no pipeline, tokenizer or architecture, and ChronoGPT-Instruct adds instruction tuning to the former.} A gain in all of them is therefore a property of the update rather than of any one release, and \textit{\new{a setting chosen for each backbone without the evaluation window} raises its IC everywhere}.

\subsection{Effect of Staleness}
\label{sec:stale}

To see whether an older checkpoint scores worse on the decision it is used for, we regress \new{each measure} on staleness \new{with suite fixed effects}, holding the evaluation window fixed. Slopes are per year of staleness, over the eligible checkpoints of Table~\ref{tab:ckpt}. As shown in Figure~\ref{fig:main}(b) and Table~\ref{tab:stale}, \new{the slope of the IC is positive, and its interval rules out a decay larger than 0.14$\times10^{-3}$ per year, under a tenth of the gain of PALM\rtwo{, although the design detects with 80\% power only a decay larger than 0.35$\times10^{-3}$ per year}. The return and the Sharpe ratio lean towards the newer vintage but stay within their intervals, and no measure shows a decay we can detect.} \new{This shows that} \textit{the more recent checkpoint is not \new{detectably} better}\new{, on every return window and without any one suite (Appendix~\ref{app:checks})}\rtwo{, and the adapted checkpoints show no decay we can detect either}.

\subsection{\new{Comparison with Full Finetuning}}
\label{sec:fullft}

\new{To check how much is lost by training only a low-rank adapter, we compare PALM with full finetuning, which trains every weight on the same eligible text (Appendix~\ref{app:alt}). As shown in Figure~\ref{fig:fullft}(a)--(c), the two updates reach similar levels on every measure, with PALM at 3.40 against 3.83 on the IC, 3.19 against 3.34 on the return and 0.66 against 0.67 on the Sharpe ratio, and the paired difference between them has an interval containing zero on each, $-0.43$ $[-1.38, +0.51]$ on the IC. Figure~\ref{fig:fullft}(d) shows the same checkpoint by checkpoint, where the gains of PALM lie close to the diagonal, slightly below it, so the checkpoints that full finetuning improves most are the ones PALM improves most. PALM thus comes close to full finetuning while training under 0.5\% of the weights and storing tens of megabytes at most per decision date instead of a full copy (Table~\ref{tab:compute}). \rtwo{Note that both updates are negligible next to the roughly 2{,}100 GPU hours of a new DatedGPT vintage (Appendix~\ref{app:cost}), so the cost that separates them is the storage each decision date requires.} Same-firm retrieval, which changes the input and not the weights, gains more, and PALM adds little to it (Appendix~\ref{app:alt}).}

\section{Analysis}

\subsection{\new{Where the Gain Comes From}}
\label{sec:where}

\begin{table}[t]
{\revcolor
\caption{\new{Where the gain comes from. \rtwo{On the 18 vintages both adapters cover,} an adapter fitted on financial text from before the cutoff matches the gain of PALM, which reads text after it, showing that \textit{the gain comes from the domain, so the eligibility rule has no detectable cost}.}}
\label{tab:domain}
\small
\setlength{\tabcolsep}{2.6pt}
\resizebox{\columnwidth}{!}{%
\begin{tabular}{l|cc|c|cc}
\toprule
Text the adapter reads & Domain & Recency & IC & IC gain [95\% CI] & Wins \\
\midrule
None (as released) & \no & \no & 0.87 & \textemdash & \textemdash \\
{[A]} Before the cutoff & \yes & \no & \best{2.67} & \best{+1.80} [-0.04, +3.56] & \best{94\%} \\
{[B]} After the cutoff & \yes & \yes & \best{2.24} & \best{+1.37} [-0.45, +3.10] & \best{89\%} \\
\bottomrule
\end{tabular}}
}
\end{table}

\new{\textbf{Two sources of the gain.} The gain of PALM can come from two sources. (1) \textit{Domain}: the adapter tunes a checkpoint, pretrained on general text that may include some financial text, to the register of financial news. (2) \textit{Recency}: the adapter reads text published after the cutoff, which the checkpoint has never seen. PALM supplies both, so to separate them we fit the same adapter, at the same rank, steps and batch, on FNSPID text published before the cutoff, in the period the vintage's pretraining already covers, rather than between the cutoff and the decision date as PALM does. This adapter supplies the domain without recency. Since FNSPID starts in 2014, the comparison runs on the \NDomain{} vintages cut from 2014 onward, and for the earliest of them the text before the cutoff spans fewer years than the text PALM reads, which works against this adapter.}

\new{\textbf{Domain, not recency.} As shown in Table~\ref{tab:domain} and Figure~\ref{fig:domain}, the domain alone recovers the gain of PALM. Across checkpoints the two gains move together, neither lies systematically above the diagonal, and their averages cannot be told apart. \rtwo{This agrees with earlier evidence that temporal adaptation gains largely reflect the domain rather than the date of text \citep{rottger2021temporal,agarwal2022temporal}.} Two consequences follow. First, the eligibility rule has no detectable cost, since an adapter that reads only text older than the checkpoint already delivers the gain, so the adapter needs no newer text. Second, it explains Section~\ref{sec:stale}. A newer vintage reads more general text, while the decision rewards the register of financial news, which is why a newer vintage is no better and a small update on financial text is.}

\begin{table}[t]
\makeatletter\def\@captype{figure}\makeatother
\centering
\includegraphics[width=\columnwidth]{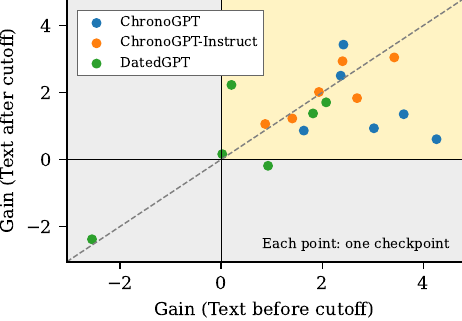}
\Description{For each checkpoint, the IC gain of an adapter fitted on text from before the cutoff against the gain of PALM. The points lie along the dashed diagonal.}
\caption{\new{Gain by domain vs. domain + recency. The x-axis is the IC gain of an adapter fitted on financial text before the cutoff (i.e., effect of domain), and the y-axis that of PALM, fitted on financial text after it (i.e., effect of domain and recency), in units of $10^{-3}$, indicating that \textit{adaptation to the financial domain drives the gain, while recency adds little}.}}
\label{fig:domain}
\vspace{10pt}
\makeatletter\def\@captype{table}\makeatother
{\revcolor
\caption{\new{Robustness of the gain. Each row changes one choice of the protocol, [1] to [5] as listed in Section~\ref{sec:robust}, showing that \textit{the gain keeps its sign whichever choice is changed}.}}
\label{tab:robust}
\small
\setlength{\tabcolsep}{2.6pt}
\resizebox{\columnwidth}{!}{%
\begin{tabular}{l|c|cc}
\toprule
Change & $n$ & IC gain [95\% CI] & Wins \\
\midrule
{[1]} Firm-level cross-section & 44 & \best{+1.56} [-0.69, +3.66] & \best{93\%} \\
{[2]} Syndicated copies dropped & 44 & \best{+2.12} [+0.09, +4.14] & \best{93\%} \\
{[3]} Padding left out of the loss & 44 & \best{+1.69} [-0.34, +3.80] & \best{84\%} \\
\multicolumn{4}{l}{[4] Random seed} \\
\quad Seed 1 & 44 & \best{+1.55} [-0.26, +3.37] & \best{89\%} \\
\quad Seed 2 & 44 & \best{+1.88} [-0.02, +3.85] & \best{93\%} \\
\quad Seed 3 & 44 & \best{+1.63} [-0.18, +3.43] & \best{91\%} \\
\multicolumn{4}{l}{[5] Decision date (shared setting, rank 2)} \\
\quad November 2016 & 32 & \best{+0.95} [-0.07, +2.00] & \best{88\%} \\
\quad November 2017 & 35 & \best{+1.40} [+0.32, +2.52] & \best{94\%} \\
\quad November 2018 & 38 & \best{+1.09} [-0.05, +2.19] & \best{89\%} \\
\quad November 2019 & 41 & \best{+1.53} [+0.30, +2.75] & \best{88\%} \\
\quad November 2021 & 47 & \best{+1.71} [+0.05, +3.28] & \best{87\%} \\
\quad November 2022 & 47 & \best{+1.47} [-0.67, +3.61] & \best{79\%} \\
\bottomrule
\end{tabular}}
}
\end{table}

\subsection{\new{Robustness of the Gain}}
\label{sec:robust}

\new{To check that the gain does not rest on a choice of protocol, we vary one choice at a time, keep the selected setting otherwise, and pair each variant with the same checkpoints as released on next-day open-to-close returns, across the five settings that follow:}
\begin{itemize}
\item \new{[1] \textbf{Firm-level cross-section}: Firms are ranked instead of articles, using the average score of a firm's articles on each day.}
\item \new{[2] \textbf{Syndicated copies dropped}: Articles that repeat another article of the same day word for word are removed.}
\item \new{[3] \textbf{Padding left out of the loss}: The adapter is not trained on the filler tokens that pad each batch.}
\item \new{[4] \textbf{Random seed}: The adapter is refitted with three other seeds.}
\item \new{[5] \textbf{Decision date}: The whole experiment is repeated with the decision date moved to each November from 2016 to 2022, without the three vintages cut in 2021 at the last date.}
\end{itemize}
\new{As shown in Table~\ref{tab:robust}, the gain keeps its sign under every change. Figure~\ref{fig:capacity} varies the rank of the adapter, with every backbone at that rank, 600 steps (fewer for a few DatedGPT vintages) and a rate of $10^{-4}$, read only after the setting was fixed. As shown in Figure~\ref{fig:capacity}, the gain stays positive at every rank, from $+1.24$ to $+2.24$ in panel (a), and the adapter raises the IC on 89 to 98\% of the checkpoints in panel (b). Appendix~\ref{app:sens} further varies the schedule, the share of text, the training budget and the length read per article, for reference only, since the setting we report is chosen on the validation window.}

\subsection{\new{Point-in-Time Audit of the Target}}
\label{sec:align}

\begin{figure}[t]
\centering
\includegraphics[width=\columnwidth]{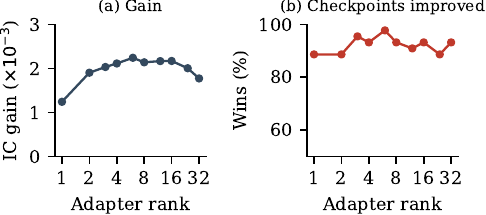}
\Description{Two panels against adapter rank on a log scale from 1 to 32. The left gives the IC gain, positive at every rank and lowest at rank 1, and the right the share of checkpoints the adapter improves, between 89 and 98 percent.}
\caption{\new{Effect of the adapter capacity. (a) The IC gain and (b) the share of checkpoints whose IC the adapter raises, at each rank, showing that \textit{the gain does not hinge on the rank}.}}
\label{fig:capacity}
\vspace{10pt}
\centering
\includegraphics[width=\columnwidth]{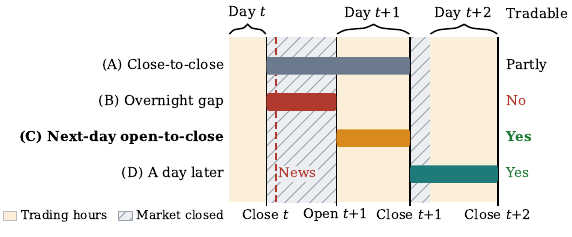}
\Description{A schematic trading timeline over three days with market hours shaded and a dashed line for news released after the close of day t. Four bars mark the return windows: close to next close, the overnight gap from the close to the next open, the next day's open to close, and the day after.}
\caption{\new{Return windows for an article dated day $t$. An article released after the close moves the price in the overnight gap (B) before anyone can trade on it, while (C) starts once every article dated $t$ is public, showing that \textit{(C) is the first window that is tradable whatever the hour of publication}.}}
\label{fig:timeline}

\end{figure}

\begin{table}[t]
{\revcolor
\caption{\new{Point-in-time audit of the target. FNSPID dates carry no time of day, so the return an article is scored against decides whether the signal could have been traded. The gain of PALM sits in (C), tradable whatever the hour of publication, and not in the overnight gap (B).}}
\label{tab:align}
\small
\setlength{\tabcolsep}{2.6pt}
\resizebox{\columnwidth}{!}{%
\begin{tabular}{l|cc|cc}
\toprule
Return scored against & Base & + PALM & IC gain [95\% CI] & Wins \\
\midrule
(A) Close $\rightarrow$ Next close & 7.70 & \best{8.81} & \best{+1.11} [-0.65, +2.99] & \best{84\%} \\
(B) Close $\rightarrow$ Next open & 14.19 & \best{14.49} & \best{+0.29} [-1.84, +2.48] & \best{55\%} \\
(C) Next open $\rightarrow$ Next close & 1.63 & \best{3.40} & \best{+1.77} [-0.15, +3.71] & \best{93\%} \\
(D) Next close $\rightarrow$ Close after & 3.19 & \best{3.33} & \best{+0.15} [-2.24, +2.34] & \best{61\%} \\
\bottomrule
\end{tabular}}
}
\vspace{10pt}
{\revcolor
\caption{\new{What the gain needs. Scrambling the word order of the eligible text removes most of the gain, while text read past the decision date adds nothing, showing that \textit{the gain needs what the text says, and nothing beyond the eligible text}.}}
\label{tab:bounds}
\small
\setlength{\tabcolsep}{3pt}
\resizebox{\columnwidth}{!}{%
\begin{tabular}{l|cc|cc|cc}
\toprule
\multirow{2}{*}[-0.25\normalbaselineskip]{Adapter text} & \multicolumn{2}{c|}{IC} & \multicolumn{2}{c|}{Return} & \multicolumn{2}{c}{Sharpe} \\
\cmidrule(lr){2-3}\cmidrule(lr){4-5}\cmidrule(lr){6-7}
 & Gain & Wins & Gain & Wins & Gain & Wins \\
\midrule
PALM & \best{+1.77} & \best{93\%} & \best{+1.27} & \best{82\%} & \best{+0.26} & \best{82\%} \\
(a) Shuffled word order & \best{+0.33} & \best{57\%} & \best{-0.57} & \best{43\%} & \best{-0.12} & \best{43\%} \\
(b) Text past the date & \best{+1.79} & \best{89\%} & \best{+1.05} & \best{82\%} & \best{+0.22} & \best{82\%} \\
\bottomrule
\end{tabular}}
}
\end{table}

\new{Look-ahead bias can enter through the target as well as through the model, when an article is scored against a price move that began before the article was public. To check that every score could have been acted on, we audit when each article becomes public. Almost every article is stamped at midnight UTC (\ShareMidnight{} of the articles in the panel), so its date is known but its hour of release is not, and an article released after the close of day $t$ is still dated $t$ \rtwo{(Figure~\ref{fig:timeline})}.}

\new{Figure~\ref{fig:timeline} lays out the returns such an article can be scored against. The usual close-to-close return (A) contains the overnight gap (B), in which the price already reacts to an article released after the close, before anyone could trade on it. The next day's open-to-close return (C) starts only after every article dated $t$ is public, so it can be traded whatever the hour of publication, and (D) checks whether the signal lasts a day longer. As shown in Table~\ref{tab:align}, the checkpoints as released read the overnight gap best, while the gain of PALM lies in (C) and fades by (D). We therefore score every result on (C), and report (A) as well, for comparison with earlier work.}

\subsection{\new{Other Analyses}}
\label{sec:other}
\label{sec:domain}

\new{\textbf{What the gain needs.} To check what the update needs, we scramble the word order of the eligible corpus and, as an upper bound no deployed system could use, let the adapter read past the decision date. As shown in Table~\ref{tab:bounds}, on the 44 checkpoints at the selected settings, scrambling removes most of the gain, and reading past the decision date does not add to it. The gain thus needs what the text says, and nothing more. Appendix~\ref{app:alt} further compares PALM with alternatives that read the same text, including full finetuning and same-firm retrieval, together with what each one costs.}

\begin{table}[t]
{\revcolor
\caption{\new{Trading costs. Turnover is two-sided and per day, and the breakeven cost is the one-way cost per unit traded at which the gross return vanishes. A daily book on article-level news does not survive realistic costs for either arm, showing that \textit{the gain is in signal quality, not in a tradable book}.}}
\label{tab:cost}
\small
\setlength{\tabcolsep}{1.9pt}
\begin{tabular}{l|c|*{4}{>{\centering\arraybackslash}p{18.5pt}}|c}
\toprule
\multirow{2}{*}[-0.25\normalbaselineskip]{Arm} & \multirow{2}{*}[-0.25\normalbaselineskip]{Turnover} & \multicolumn{4}{c|}{Sharpe after one-way cost} & \multirow{2}{*}[-0.25\normalbaselineskip]{\begin{tabular}[c]{@{}c@{}}Breakeven\\(bp)\end{tabular}} \\
\cmidrule(lr){3-6}
 & & 0 bp & 1 bp & 2 bp & 5 bp & \\
\midrule
Base & 1.62 & 0.40 & -0.47 & -1.34 & -3.94 & 0.47 \\
+ PALM & 1.62 & 0.66 & -0.19 & -1.04 & -3.60 & 0.78 \\
\midrule
Gain & 0.00 & \best{+0.26} & \best{+0.28} & \best{+0.29} & \best{+0.34} & \best{+0.31} \\
\bottomrule
\end{tabular}
}
\end{table}

\new{\textbf{Turnover and trading costs.}\label{sec:cost} To check whether the gain survives trading, we measure the daily turnover of the long-short book and the Sharpe ratio after a one-way cost. As shown in Table~\ref{tab:cost}, the book rebuilds most of its positions every day, so a cost of a single basis point already removes its gross Sharpe ratio for both arms, and PALM moves the breakeven cost by a fraction of a basis point. We therefore do not read the portfolio measures as tradable returns. This shows that the gain lies in the quality of the signal rather than in a tradable book, and we therefore treat the IC as the main measure of what PALM changes.}

\begin{table}[t]
{\revcolor
\caption{\new{Robustness to the direction words.} The same comparison \new{is} read through three disjoint sets of direction words\rtwo{, on the close-to-close target at the shared setting and on 19 checkpoints}. The unadapted score moves with the wording while the adapted score does not, showing that \textit{the gain belongs to the update and not to the words that read it out}.}
\label{tab:words}
\small
\setlength{\tabcolsep}{4pt}
\begin{fitw}{\columnwidth}
\begin{tabular}{l|cc|c|c}
\toprule
\multirow{2}{*}[-0.25\normalbaselineskip]{Wording} & \multicolumn{2}{c|}{IC} & \multirow{2}{*}[-0.25\normalbaselineskip]{IC gain} & \multirow{2}{*}[-0.25\normalbaselineskip]{Wins} \\
\cmidrule(lr){2-3}
 & Base & + PALM & & \\
\midrule
Reported & 6.65 & \best{8.32} & \best{+1.67} & \best{95\%} \\
\quad{\footnotesize $\mathcal{W}^{+}$: rise, increase, go up, gain} & & & & \\
\quad{\footnotesize $\mathcal{W}^{-}$: fall, decrease, go down, drop} & & & & \\
\midrule
Alternative A & 5.23 & \best{8.27} & \best{+3.04} & \best{89\%} \\
\quad{\footnotesize $\mathcal{W}^{+}$: climb, advance, strengthen, rally} & & & & \\
\quad{\footnotesize $\mathcal{W}^{-}$: slide, retreat, weaken, tumble} & & & & \\
\midrule
Alternative B & 4.81 & \best{8.27} & \best{+3.46} & \best{89\%} \\
\quad{\footnotesize $\mathcal{W}^{+}$: up, higher, better, stronger} & & & & \\
\quad{\footnotesize $\mathcal{W}^{-}$: down, lower, worse, weaker} & & & & \\
\bottomrule
\end{tabular}
\end{fitw}
}
\end{table}

\begin{table}[t]
{\revcolor
\caption{\new{A worked example.} One scored date, 13 December 2021, \new{is shown} for the ChronoGPT vintage cut at the end of 2019, and an article tagged with several tickers enters once for each. Signals are compared within the date, and the individual outcomes often disagree with the ordering, showing that \textit{the signal lives in the average over many dates rather than in any one of them}.}
\label{tab:worked}
\small
\setlength{\tabcolsep}{2.5pt}
\begin{tabular}{l|rr|l}
\toprule
Ticker & $f_\theta$ & Return & Headline \\
\midrule
ABT  & $+1.53$ & $-2.22\%$ & Abbott rewards shareholders as shares peak \\
INCY & $+1.42$ & $+1.15\%$ & Incyte reports a survival benefit \\
VALE & $+1.36$ & $-0.73\%$ & Nucor acquires a majority stake \\
\midrule
SPWR & $-1.22$ & $-10.78\%$ & California to reform solar policy \\
PCG  & $-1.27$ & $-0.90\%$ & California to reform solar policy \\
AAL  & $-1.27$ & $-1.23\%$ & Why airline stocks are falling \\
\bottomrule
\end{tabular}
}
\end{table}

\new{\textbf{Choice of direction words.} To check that the gain does not hinge on the words that read the signal out, we repeat the comparison with two disjoint sets of direction words, on the close-to-close target at the shared setting of rank 2, on the 19 checkpoints for which all three wordings were run. As shown in Table~\ref{tab:words}, the base checkpoint moves with the wording, from 4.81 to 6.65, while the adapted checkpoint stays within 0.05 of the same value, so the gain holds under every wording and is smallest under the one we report.}

\new{\textbf{A worked example.} To show what one score means, we follow one scored date, 13 December 2021, for the ChronoGPT vintage cut at the end of 2019, on the close-to-close target, with every stage in Appendix~\ref{app:example}. As shown in Table~\ref{tab:worked}, articles are ranked within the date, and the returns often disagree with the ordering, so the signal lives in the average over many dates, not in any one.}

\section{Limitations}
\label{sec:limitations}

Our evidence comes from one asset class over a single decade, and from the point-in-time suites that are public today. Whether the same picture holds for other markets, other prediction horizons, or backbones an order of magnitude larger is beyond what a study of this scope settles. The signal we measure is weak in absolute terms, as signals extracted from news generally are, and conclusions about stronger signals, or about tasks that need knowledge of recent events, do not follow from it.

\new{Four limits are specific to this version. First, the staleness test rules out only a large yearly decay, and no eligible vintage is less than a year stale, so smaller or earlier decays would need a study with more vintages per suite. Second, the panel carries no time of day, which is why we use next-day open-to-close returns rather than intraday ones. Third, the long-short book does not survive realistic trading costs for either arm, and the universe favours firms that survived to the data release, so the gain we report is in the quality of a ranking signal, not in a tradable strategy. Fourth, the gain is smallest on DatedGPT and changes sign across its decision dates, so the robustness we report holds for the pooled gain, and the adapter does not help the larger PIT-4B (Appendix~\ref{app:pit4b}).}

\section{Conclusion}

In this paper, we tested whether the annual pretraining run of a PIT suite is necessary. Holding the evaluation window fixed, and comparing each vintage against the vintage that replaced it \new{on a target that a trader could have acted on}, we found that staleness does not degrade downstream performance \new{by any amount we can detect}. Motivated by this observation, we proposed PALM, a simple yet effective plug-in method that fits a low-rank adapter on eligible text without modifying any pretrained weight\new{, with its setting chosen on a separate validation window}.

Our results suggest that a PIT suite should be maintained by \textit{updating the checkpoint it already has} rather than by producing a new one each calendar year. \new{Since an adapter fitted on text older than the checkpoint recovers the same gain, what a stale checkpoint lacks is the register of financial news rather than the news of the years it missed, and the point-in-time constraint has no detectable cost.} They also suggest that the update can be small, since an adapter of a few directions already recovers the gain. This preserves the chronological guarantee the suite is built to provide, and frees the budget the calendar consumes for the coverage that is actually missing, which is more architectures\new{, not} more years of one.

\clearpage
\bibliographystyle{ACM-Reference-Format}
\bibliography{refs}

\clearpage
\onecolumn
\appendix
\section{Experimental Setup Details}
\label{app:setupdetails}

\subsection{Dataset}
\label{app:data}

\new{\noindent\textbf{Backtest specification.} The cross-section is the article (\ArticlesPerDay{} articles over \TickersPerDay{} distinct tickers on the median day), and a date enters only with at least eight articles and four distinct scores. A ticker carries more than one article on \ShareMultiArticle{} of ticker-days, and \ShareDupText{} of articles repeat another text of the same day verbatim. Both are handled in Section~\ref{sec:robust}, where the cross-section is moved to the firm and syndicated copies are dropped. The universe is the set of tickers with a price history in the FNSPID release, which favours firms that survived to it, as a property of the panel shared by every arm. No text of the evaluation window enters any adapter, as the latest date each adapted panel read confirms, except the upper bound of Table~\ref{tab:bounds}(b), which reads past it by design. Returns are gross, and Section~\ref{sec:cost} prices trading costs.}

The panel is built from FNSPID \citep{dong2024fnspid}, a public corpus of dated financial news. We sample it evenly across months, which keeps any single year from dominating the pool. Each row is one article, and a ticker carrying several articles on one day contributes several rows rather than one average, which is why Section~\ref{sec:setup} counts the panel in article-days rather than in firm-days of the kind a daily study would use.

An article joins a trading day only when its timestamp precedes that day's close, taken conservatively at 20:00 UTC. A story stamped after the close therefore belongs to the next trading day, while one stamped only by its date stays on that date (Section~\ref{sec:align}). \new{The target is the open-to-close return of the \textit{following} session (Section~\ref{sec:align}), and never a return that starts before the article could have been read.} The decision date then splits the panel into the two halves the method needs, one an adapter may read and one every checkpoint is scored on. Table~\ref{tab:data} \new{in Section~\ref{sec:setup}} gives the size of each half and the rest of what the panel holds.

\subsection{Backbones}
\label{app:models}

Four public point-in-time suites are used, and they divide into \new{three} kinds.
\begin{itemize}
\item \new{DatedGPT \citep{yan2026datedgpt} trains each vintage from scratch, so a vintage is a self-contained model.}
\item \new{ChronoGPT \citep{he2025chrono} trains its first vintage on text through 1999 and each later one by continuing the previous vintage on the next year's text, and ChronoGPT-Instruct \citep{he2025instruct} instruction-tunes the ChronoGPT vintage of the same cutoff. No vintage has read text after its cutoff.}
\item PIT-4B \citep{kelly2026pit} publishes monthly snapshots of one continuous run, \new{wider than the others, and we report it separately in Appendix~\ref{app:pit4b}.}
\end{itemize}

Every checkpoint is loaded at the settings of its original release, in bfloat16 and in evaluation mode, and nothing about the architecture is changed. One scoring routine reads all four suites, which keeps the comparison between them free of any difference in how a score was taken. Table~\ref{tab:backbones} lists what each suite releases and how much of it is eligible for our decision date.

\begin{table}[H]
\caption{The four suites. The three suites we pair release one vintage per calendar year without gaps and their eligible counts differ only through where each suite starts, while PIT-4B publishes monthly checkpoints from which we take one per calendar year, showing that \textit{the paired pool is set by release history rather than by any choice of ours}.}
\label{tab:backbones}
\begin{fitw}{\columnwidth}
\begin{tabular}{l|l|c|cc}
\toprule
\multirow{2}{*}[-0.25\normalbaselineskip]{Suite} & \multirow{2}{*}[-0.25\normalbaselineskip]{Hugging Face identifier} & \multirow{2}{*}[-0.25\normalbaselineskip]{Width} & \multicolumn{2}{c}{Vintages} \\
\cmidrule(lr){4-5}
 & & & Released & Eligible \\
\midrule
ChronoGPT & \texttt{manelalab/chrono-gpt-v1-*} & 1536 & 26 (1999--2024) & 21 (1999--2019) \\
ChronoGPT-Instruct & \texttt{manelalab/chrono-gpt-instruct-v1-*} & 1536 & 21 (2004--2024) & 16 (2004--2019) \\
DatedGPT & \texttt{datedgpt/datedgpt-*-base} & 2048 & 12 (2013--2024) & 7 (2013--2019) \\
\midrule
PIT-4B & \texttt{Diamegs/PIT-4B-*} & 4096 & Monthly (2013--2024) & 7 used (2013--2019) \\
\bottomrule
\end{tabular}
\end{fitw}
\end{table}

\clearpage
\section{Three Reported Measures}
\label{app:metrics}

\new{The worked numbers here follow Section~\ref{app:example}, which uses the earlier close-to-close target, and the definitions hold for any target.}

All three measures are read off one daily series. Each trading day of the evaluation window yields a cross-section of scores and the returns those scores were meant to order, and contributes one number to each of the three. They differ in what they do with a day rather than in the data they read, which is why a variant can move one of them while leaving the other two where it found them.

The unit of the cross-section is the \textit{article} and not the firm, since a ticker carrying several articles on one day enters the ranking several times with the same realized return, and the median day carries 99 articles over 86 distinct tickers. A day is scored only when it holds at least eight articles and four distinct scores, which removes 6 of the 773 dates in the window. Both conventions hold everywhere in the paper.

The tables report these measures in fixed units. The IC is given in units of $10^{-3}$ and the annualized return in percentage points, and Wins is the share of paired checkpoints a variant improves. Bold marks the better entry of a pair, and the gain and its share of improved checkpoints in the tables of adapted arms.

\subsection{Information Coefficient}

The information coefficient asks whether the ordering the model produced on a given day matched the ordering the market delivered. Writing $\rho$ for the Spearman rank correlation, $f_\theta(x_{jt})$ for the score of article $j$ on day $t$, and $r_{j,t+1}$ for \new{the next-day open-to-close return}, Eq.~\eqref{eq:ic} averages the daily correlation over the window $T$,
\begin{equation*}
\mathrm{IC} = \frac{1}{|T|} \sum_{t \in T} \rho \big( \{ f_\theta(x_{jt}) \}_j , \; \{ r_{j,t+1} \}_j \big) .
\end{equation*}
The correlation is computed \textit{within} a day, which removes the move common to every name that session and prevents a market direction from being read as a signal. It is a \textit{rank} correlation, which makes the measure invariant to the units of $f_\theta$ and keeps a single large return from setting the value on its own. The daily correlations are then averaged with equal weight, which keeps a day with many articles from counting for more than a thin one. Each of the three choices keeps the measure from reading something other than the ordering it is meant to score.

On the worked example of Section~\ref{app:example}, the 91 articles of 13 December 2021 give $\mathrm{IC}_t = +0.0067$, and averaging over the 767 scored days gives $0.00661$ for that checkpoint, \new{printed as $6.61$ in units of $10^{-3}$ on that target}.

\subsection{Annualized Return}

The information coefficient scores an ordering, and the return scores what an ordering is worth once positions are taken. The score is demeaned within the day, using that day's mean score $\bar f_t$, and its sign becomes the position, which gives a long-short book that is equally weighted by construction,
\begin{equation*}
w_{jt} = \operatorname{sign}\big(f_\theta(x_{jt}) - \bar f_t\big), \qquad
r_t = \frac{\sum_j w_{jt}\, r_{j,t+1}}{\sum_j |w_{jt}|} , \qquad
\text{Return} = 252\,\bar r .
\end{equation*}
Taking the sign rather than the score itself is what separates this measure from the IC. The IC uses the whole ordering, while the book only asks which side of the day's average an article fell on, and it therefore discards the magnitude of the score. That is why an adapter can raise the IC while leaving the book unmoved, and \new{Table~\ref{tab:allck} shows cases where the two disagree in sign}.

The book is not exactly dollar neutral. The sign of a demeaned score splits a day into two sides, but the split need not be even, and on 13 December 2021 it was 55 long against 36 short. The denominator $\sum_j |w_{jt}|$ normalizes by the number of positions rather than by the smaller side, and the reported return is therefore the average return of a position taken that day. No transaction cost, borrowing cost or slippage is deducted, which makes every return here a gross figure to be read as a difference between arms rather than as a tradable one.

For the same checkpoint, $r_t = -0.33\%$ on the example day, and $252\,\bar r = 9.19\%$ over the window. That annualization multiplies a daily mean by 252 and does not compound, which keeps the measure additive across arms and lets the difference between two rows be read directly as a difference in annualized terms rather than in daily ones.

\subsection{Sharpe Ratio}

The Sharpe ratio divides the same daily series by its own volatility, which is what a desk is held to rather than the size of the return,
\begin{equation*}
\text{Sharpe} = \sqrt{252} \; \frac{\bar r}{\sigma_r} ,
\end{equation*}
with $\bar r$ and $\sigma_r$ the mean and the standard deviation of $r_t$ over the window. No risk-free rate is subtracted, which is standard for a long-short book that holds no net cash position, and the $\sqrt{252}$ scales a daily ratio to an annual one under the usual assumption that the daily returns carry no serial correlation from one session to the next.

The Sharpe ratio and the return move together but not identically, since a variant can raise the mean while raising the dispersion by more. For the checkpoint above, a daily mean of $0.0004$ against a daily standard deviation of $0.0035$ gives $1.65$. The tables of Section~\ref{app:sens} therefore report the three together, since a change that lifts all three at once is \textit{a different kind of evidence} from one that lifts only the IC.

\clearpage
\section{A Worked Example of the Pipeline}
\label{app:example}

\new{The example uses the close-to-close target, and only the return in Stage 4 changes on next-day open-to-close returns.}

This section follows one article through every stage of Section~\ref{sec:setup}, from the string the model reads to the numbers the tables report. The checkpoint is the ChronoGPT vintage cut at the end of 2019, which is eligible for the decision date at the end of November 2020, and the article is drawn from 13 December 2021, inside the evaluation window.

\textbf{Stage 1: The prompt.} Scoring uses one template throughout. The ticker and the article text are substituted into it, the text is truncated at 1{,}500 characters, and the string stops immediately before the word that would name the direction.

\begin{quote}\ttfamily\small\raggedright
News about AAL: Why Airline Stocks Are Falling Today. Shares of American Airlines Group (NASDAQ: AAL), United Airlines Holdings (NASDAQ: UAL), Delta Air Lines (NYSE: DAL), and Southwest Airlines (NYSE: LUV) all fell as much as 5\% on Monday, a weak day for markets overall. $\ldots$

Following this news, the stock price of AAL will
\end{quote}

The publication date never enters the prompt. It is used outside the model to assign the article to a trading day, to define the cross-section it is ranked within, and to select the return it is scored against. The prompt itself therefore carries \textit{no marker of when the article was written}.

\textbf{Stage 2: The direction words.} The next-token distribution at the final position is read once, and the probability mass on two disjoint sets of continuations is compared. The reported results use
\begin{align*}
\mathcal{W}^{+} &= \{\text{\ttfamily\ rise},\ \text{\ttfamily\ increase},\ \text{\ttfamily\ go up},\ \text{\ttfamily\ gain}\}, \\
\mathcal{W}^{-} &= \{\text{\ttfamily\ fall},\ \text{\ttfamily\ decrease},\ \text{\ttfamily\ go down},\ \text{\ttfamily\ drop}\}.
\end{align*}
Each set is compared at the first token position at which its members differ, since a tokenizer that emits the leading space separately would otherwise return the same identifier for every candidate and collapse the contrast to zero. Substituting Eq.~\eqref{eq:score} gives $f_\theta = -1.272$ for the article above, a negative value that places it at the bottom of its date.

\textbf{Stage 3: The cross-section.} All 91 articles dated 13 December 2021 are ranked by $f_\theta$. \new{Table~\ref{tab:worked} in Section~\ref{sec:other}} lists the three highest and the three lowest. The model is right about the airline story and wrong about the dividend, which is ordinary at this signal strength.

\textbf{Stage 4: The two daily numbers.} The Spearman correlation between the 91 scores and the 91 realized \new{close-to-close} returns is $\mathrm{IC}_t = +0.0067$. Taking the sign of the score demeaned within the date and holding each name at equal weight gives a long-short return of $r_t = -0.33\%$ on that day. Averaging $\mathrm{IC}_t$ over the 767 scored days of the window gives $6.61$ in units of $10^{-3}$ for this checkpoint, and annualizing $r_t$ gives the return and the Sharpe ratio of the same row.

\textbf{Stage 5: What PALM changes.} Nothing above involves a return label or any trained parameter. PALM leaves Stages 1 to 4 untouched and changes only the weights that produce $f_\theta$, by fitting the adapter of Eq.~\eqref{eq:palm} on the text published between the end of 2019 and the decision date. The same 91 articles are then scored again by the adapted checkpoint, and the difference between the two rows of every table is the difference between those two sets of scores.

\clearpage
\section{Implementation and Setting Selection}
\label{app:implsel}

\subsection{Implementation Details}
\label{app:impl}

Every pretrained parameter is frozen before a single adapter is created, since injecting a module does not by itself freeze the layers it was not placed in, and leaving them trainable would make the run a full finetune rather than a low-rank update. The suites do not name their layers alike, and the loader matches one shared list of names, which selects in every block the attention output and both feed-forward projections of ChronoGPT, and the query, value and up projections of DatedGPT. Initializing $B$ at zero leaves the adapted model identical to the vintage it was attached to until the first optimizer step is taken.

The optimizer is AdamW \citep{loshchilov2019adamw}, and each batch is drawn with replacement from the eligible interval. \new{The rank, the steps and the rate are chosen on the validation window only (Appendix~\ref{app:select}), and Section~\ref{app:sens} reports the evaluation window around the budget for reference.}

Reading a score involves no trained parameter and no sampling. The same input therefore returns the same score on every run. Table~\ref{tab:impl} collects the settings that hold everywhere in the paper unless a sensitivity arm states otherwise.

\new{\textbf{Measuring the cost.}\label{app:cost} Table~\ref{tab:compute} is measured on one NVIDIA L40S with no other process on the GPU, which we checked once a second during every run. Each update is run on the DatedGPT vintage cut at the end of 2019, at the setting it uses in the paper, PALM at rank 16 for 300 steps at $10^{-4}$ and full finetuning on every weight for 600 steps at $10^{-5}$, both with a batch of 2 and the same eligible text. Training time runs from the moment the model is loaded and the trainable weights are set to the last update, so loading and scoring are left out. Full finetuning is timed over the 300 steps of PALM and scaled to its own 600, since every step reads a batch of the same size. Tokens read are an upper bound of steps $\times$ batch $\times$ 256, since shorter articles are not padded to that length. Peak memory is read from the GPU over the whole job, whose scoring part is the same for both updates, and storage counts the weights kept for one decision date, in bfloat16 for a model and in float32 for an adapter. For a new vintage, we measure full-parameter language modeling on packed sequences of 2048 tokens at the largest batch that fits, 13{,}055 tokens per second, and multiply by the 100B tokens a DatedGPT vintage reads \citep{yan2026datedgpt}, which gives 2{,}128 GPU hours, close to the 2{,}000 A100 hours its authors report. The same measurement on ChronoGPT gives 2.9 minutes for PALM and 3.8 minutes for full finetuning, against 169 GPU hours, at sequences of 1792 tokens, for the 2.6B tokens of an average yearly update.}

\begin{table}[H]
\caption{The settings that hold throughout. Everything except the rank, the steps and the rate is fixed once and never tuned per suite\new{, and those three are chosen on the validation window}, showing that \textit{the gain is not the product of a search}.}
\label{tab:impl}
\begin{fitw}{\columnwidth}
\begin{tabular}{l|l}
\toprule
Setting & Value \\
\midrule
\multicolumn{2}{l}{\textit{The adapter}} \\
\cmidrule(lr){1-2}
Rank and scale & \new{Selected per backbone (Section~\ref{sec:setup})}, $\alpha = 2r$ \\
Placement & Attention output and feed-forward projections (ChronoGPT) \\
 & Query, value and up projections (DatedGPT) \\
Initialization & $A$ random \\
 & $B = 0$ \\
 & Every pretrained weight frozen \\
\midrule
\multicolumn{2}{l}{\textit{Fitting the adapter}} \\
\cmidrule(lr){1-2}
Objective & Language modeling on $\mathcal{D}_{(t_c,t_d]}$, no return label \\
Optimizer & AdamW, no weight decay, one-cycle with a tenth warmup \\
Budget & \new{Selected per backbone}, sequences of 256 tokens \\
Batch size & 4 for the 1536-wide suites, 2 for DatedGPT \\
\midrule
\multicolumn{2}{l}{\textit{Reading a score}} \\
\cmidrule(lr){1-2}
Prompt & Template of Section~\ref{app:example}, cut at 1{,}500 characters \\
Encoding & 768 tokens (right padding) \\
Decoding & One forward pass (no sampling and no generation) \\
\bottomrule
\end{tabular}
\end{fitw}
\end{table}

\subsection{Selection of the Adapter Setting}
\label{app:select}

\begin{table}[H]
{\revcolor
\caption{\new{Selection of the adapter setting. Each cell is the pooled IC gain of one setting of the grid fixed in advance, on that backbone's validation checkpoints (\NValChrono{}, \NValInstr{} and \NValDated{} of them), and the bold cell is the setting the rule selects.}}
\label{tab:select}
\small
\setlength{\tabcolsep}{4pt}
\begin{fitw}{\columnwidth}
\begin{tabular}{c|c|ccc}
\toprule
Rank & Steps & ChronoGPT & ChronoGPT-Instruct & DatedGPT \\
\midrule
1 & 300 & -0.14 & +0.63 & -1.61 \\
1 & 600 & +0.24 & \best{+1.68} & -1.82 \\
1 & 1200 & +0.09 & +0.22 & -1.27 \\
2 & 300 & -0.15 & +1.13 & -2.03 \\
2 & 600 & -0.13 & +1.10 & -1.45 \\
2 & 1200 & -0.12 & +0.11 & -1.26 \\
4 & 300 & -0.06 & +1.39 & -1.59 \\
4 & 600 & +0.08 & +0.67 & -1.06 \\
4 & 1200 & \best{+0.77} & -0.79 & -1.40 \\
8 & 300 & -0.21 & +0.97 & -0.75 \\
8 & 600 & -0.11 & -0.34 & -0.99 \\
8 & 1200 & +0.69 & -1.91 & -1.25 \\
16 & 300 & -0.32 & +0.17 & \best{-0.20} \\
16 & 600 & +0.36 & -0.85 & -0.93 \\
16 & 1200 & -0.13 & -2.69 & -1.13 \\
\midrule
2 & 600, $5\times10^{-5}$ & -0.01 & +1.01 & -1.45 \\
2 & 600, $2\times10^{-4}$ & +0.21 & -0.04 & -0.78 \\
4 & 600, $5\times10^{-5}$ & +0.09 & +0.99 & -1.43 \\
4 & 600, $2\times10^{-4}$ & +0.64 & -1.32 & -0.89 \\
8 & 600, $5\times10^{-5}$ & +0.04 & +0.96 & -1.33 \\
8 & 600, $2\times10^{-4}$ & +0.77 & -2.44 & -0.84 \\
\bottomrule
\end{tabular}
\end{fitw}
}
\end{table}

\new{To show how the settings of Section~\ref{sec:setup} were reached, we report every setting of the grid on the validation window, for each backbone. The rule takes, for each backbone, the setting with the largest validation gain, treating gains within 0.05 of it as ties and breaking them towards the smaller rank and the fewer steps, and the evaluation window takes no part in it. As shown in Table~\ref{tab:select}, the backbones prefer different capacities, which is expected of models whose widths and training pipelines differ. Note that the validation pool holds only two DatedGPT vintages, so its choice rests on fewer checkpoints, and every setting loses on them, so the rule takes the least negative.}

\clearpage
\section{Comparison with Alternatives}
\label{app:alt}

\begin{figure}[H]
\centering
\includegraphics[width=\ifdefined\APPONECOL 0.55\textwidth\else\columnwidth\fi]{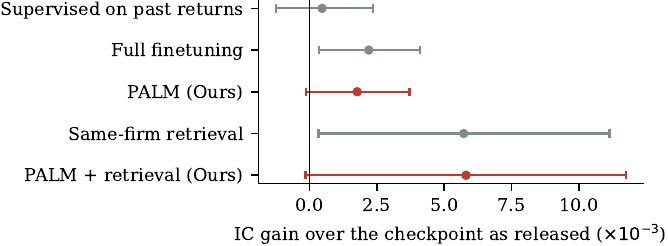}
\Description{IC gain and its 95 percent interval for supervised training, full finetuning, PALM, same-firm retrieval and PALM combined with retrieval.}
\caption{\new{Comparison with alternatives. Each bar is the IC gain and 95\% interval of an alternative of Table~\ref{tab:alt} over the base.}}
\label{fig:alt}
\end{figure}

\new{To place PALM among the alternatives a practitioner could use under the same point-in-time rule, we compare it with methods that read the same eligible text or need no checkpoint at all. Full finetuning trains every weight on the same text for 600 steps, supervised training fits the same readout to the sign of close-to-close returns realized before the decision date, the Loughran--McDonald lexicon \citep{loughran2011liability} needs no model, and same-firm retrieval prepends the five most recent articles about the same firm dated before the scored one. As shown in Table~\ref{tab:alt} and Figure~\ref{fig:alt}, PALM improves on the lexicon and on supervised training without reading a single label. Full finetuning moves the IC further, but, like the annual pretraining run, it stores a full copy of the model for every decision date, while PALM stores an adapter of under 25 MB. Retrieval changes the input rather than the weights, and needs a dated index and a longer prompt. We attribute the small increment of PALM on top of retrieval to the finding of Section~\ref{sec:where}, as retrieved articles already bring financial news into the prompt. The two therefore serve different settings, retrieval when an index can be maintained and queried, and PALM when the model must carry the domain.}

\begin{table}[H]
{\revcolor
\caption{\new{Comparison with alternatives. All \NAltCkpt{} eligible checkpoints are read on next-day open-to-close returns, with what each method costs. Every update to the weights reads the same eligible text, full finetuning and supervised training for 600 steps on every backbone and PALM at its selected setting, and retrieval adds the five latest same-firm articles dated before the scored one. PALM is the only update of the weights that needs neither labels, nor a copy of the model per decision date, nor a longer input, showing that \textit{it buys the gain at the lowest cost}.}}
\label{tab:alt}
\small
\setlength{\tabcolsep}{3.5pt}
\begin{tabular}{l|c|c|c|c|cc}
\toprule
Method & Trainable & Per decision date & Added input & IC & IC gain [95\% CI] & Wins \\
\midrule
Base & \textemdash & \textemdash & \textemdash & 1.63 & \textemdash & \textemdash \\
Loughran--McDonald lexicon & \textemdash & \textemdash & \textemdash & 1.28 & \textemdash & \textemdash \\
Supervised on past returns & 0.06--0.10\% & Under 8 MB & Return labels & 2.10 & +0.47 [-1.25, +2.34] & 52\% \\
Full finetuning & 100\% & Full copy, 2.7--3.7 GB & \textemdash & 3.83 & +2.20 [+0.36, +4.09] & 93\% \\
\best{PALM (Ours)} & 0.05--0.45\% & Under 25 MB & \textemdash & 3.40 & +1.77 [-0.15, +3.71] & 93\% \\
Same-firm retrieval & \textemdash & Dated index & Up to 900 characters & 7.35 & +5.72 [+0.33, +11.14] & 100\% \\
\best{PALM + retrieval (Ours)} & 0.05--0.45\% & Under 25 MB + index & Up to 900 characters & 7.44 & +5.81 [-0.15, +11.75] & 100\% \\
\bottomrule
\end{tabular}
}
\end{table}

\clearpage
\section{Additional Analyses at the Shared Setting}
\label{app:earlier}

\subsection{Sensitivity of the Update}
\label{app:sens}

\new{This subsection varies one choice at a time around the shared setting of rank 2, on next-day open-to-close returns and against the same checkpoints as released. These sweeps read the evaluation window, so they describe the surface around a setting and play no part in choosing the reported one, and the seeds and the decision date are in Table~\ref{tab:robust}. Reading the eligible text in one pass beats a yearly walk (Table~\ref{tab:sens-sched}), a small share of the text nearest the decision date already carries most of the gain \rtwo{(Table~\ref{tab:sens-frac})}, the update has to be trained past a floor, beyond which the IC keeps its gain while the portfolio measures fade \rtwo{(Table~\ref{tab:sens-budget})}, and the length read per article matters little (Table~\ref{tab:sens-len}).}

\begin{table}[H]
{\revcolor
\caption{\new{Adaptation schedule. Reading the eligible interval once beats walking it year by year.}}
\label{tab:sens-sched}
\small
\begin{fitw}{\columnwidth}
\begin{tabular}{l|cc|cc|cc}
\toprule
\multirow{2}{*}[-0.25\normalbaselineskip]{Schedule} & \multicolumn{2}{c|}{IC} & \multicolumn{2}{c|}{Return} & \multicolumn{2}{c}{Sharpe} \\
\cmidrule(lr){2-3}\cmidrule(lr){4-5}\cmidrule(lr){6-7}
 & Gain [95\% CI] & Wins & Gain [95\% CI] & Wins & Gain [95\% CI] & Wins \\
\midrule
One pass, whole window & \best{+1.90} [+0.48, +3.37] & \best{89\%} & \best{+1.50} [+0.19, +2.82] & \best{84\%} & \best{+0.31} [+0.03, +0.58] & \best{84\%} \\
One pass per year & \best{+1.11} [-0.16, +2.40] & \best{77\%} & \best{+1.15} [+0.08, +2.25] & \best{84\%} & \best{+0.23} [+0.01, +0.47] & \best{80\%} \\
\bottomrule
\end{tabular}
\end{fitw}
}
\end{table}

\begin{table}[H]
{\revcolor
\caption{\new{Share of the eligible text. The slice nearest the decision date is kept, and a small share already carries most of the gain.}}
\label{tab:sens-frac}
\small
\begin{fitw}{\columnwidth}
\begin{tabular}{l|cc|cc|cc}
\toprule
\multirow{2}{*}[-0.25\normalbaselineskip]{Share} & \multicolumn{2}{c|}{IC} & \multicolumn{2}{c|}{Return} & \multicolumn{2}{c}{Sharpe} \\
\cmidrule(lr){2-3}\cmidrule(lr){4-5}\cmidrule(lr){6-7}
 & Gain [95\% CI] & Wins & Gain [95\% CI] & Wins & Gain [95\% CI] & Wins \\
\midrule
5\% & \best{+1.36} [-0.14, +2.94] & \best{90\%} & \best{+1.55} [+0.12, +3.01] & \best{83\%} & \best{+0.31} [+0.01, +0.60] & \best{83\%} \\
10\% & \best{+1.52} [+0.05, +3.11] & \best{93\%} & \best{+1.35} [-0.01, +2.76] & \best{86\%} & \best{+0.27} [-0.01, +0.56] & \best{84\%} \\
25\% & \best{+1.69} [+0.30, +3.16] & \best{95\%} & \best{+1.18} [-0.19, +2.54] & \best{77\%} & \best{+0.24} [-0.05, +0.51] & \best{77\%} \\
50\% & \best{+1.65} [+0.19, +3.10] & \best{95\%} & \best{+1.29} [-0.10, +2.65] & \best{84\%} & \best{+0.27} [-0.02, +0.54] & \best{84\%} \\
75\% & \best{+1.64} [+0.16, +3.16] & \best{91\%} & \best{+1.31} [-0.04, +2.67] & \best{80\%} & \best{+0.27} [-0.02, +0.55] & \best{80\%} \\
100\% & \best{+1.90} [+0.48, +3.37] & \best{89\%} & \best{+1.50} [+0.19, +2.82] & \best{84\%} & \best{+0.31} [+0.03, +0.58] & \best{84\%} \\
\bottomrule
\end{tabular}
\end{fitw}
}
\end{table}

\begin{table}[H]
{\revcolor
\caption{\new{Training budget. Steps vary at a rate of $10^{-4}$, and the rate at 600 steps. The gain needs a floor of training, beyond which the IC holds and the portfolio measures fade.}}
\label{tab:sens-budget}
\small
\begin{fitw}{\columnwidth}
\begin{tabular}{l|cc|cc|cc}
\toprule
\multirow{2}{*}[-0.25\normalbaselineskip]{Budget} & \multicolumn{2}{c|}{IC} & \multicolumn{2}{c|}{Return} & \multicolumn{2}{c}{Sharpe} \\
\cmidrule(lr){2-3}\cmidrule(lr){4-5}\cmidrule(lr){6-7}
 & Gain [95\% CI] & Wins & Gain [95\% CI] & Wins & Gain [95\% CI] & Wins \\
\midrule
100 steps & \best{+0.03} [-0.26, +0.33] & \best{50\%} & \best{-0.01} [-0.45, +0.41] & \best{52\%} & \best{-0.00} [-0.10, +0.08] & \best{50\%} \\
200 steps & \best{+0.27} [-0.54, +1.04] & \best{61\%} & \best{+0.63} [-0.06, +1.33] & \best{73\%} & \best{+0.13} [-0.02, +0.27] & \best{70\%} \\
300 steps & \best{+0.62} [-0.65, +1.85] & \best{73\%} & \best{+1.08} [+0.05, +2.11] & \best{80\%} & \best{+0.22} [+0.00, +0.43] & \best{80\%} \\
600 steps & \best{+1.90} [+0.48, +3.37] & \best{89\%} & \best{+1.50} [+0.19, +2.82] & \best{84\%} & \best{+0.31} [+0.03, +0.58] & \best{84\%} \\
900 steps & \best{+2.04} [-0.03, +4.23] & \best{93\%} & \best{+1.64} [-0.13, +3.26] & \best{86\%} & \best{+0.34} [-0.03, +0.67] & \best{82\%} \\
1200 steps & \best{+2.21} [-0.05, +4.72] & \best{91\%} & \best{+1.39} [-0.55, +3.28] & \best{77\%} & \best{+0.28} [-0.12, +0.68] & \best{77\%} \\
1800 steps & \best{+2.23} [-0.37, +5.09] & \best{91\%} & \best{+0.93} [-1.24, +3.13] & \best{73\%} & \best{+0.19} [-0.27, +0.65] & \best{73\%} \\
2400 steps & \best{+2.13} [-0.65, +5.12] & \best{89\%} & \best{+0.66} [-1.76, +3.09] & \best{68\%} & \best{+0.13} [-0.37, +0.64] & \best{66\%} \\
Rate $5\times10^{-6}$ & \best{-0.04} [-0.29, +0.20] & \best{39\%} & \best{-0.13} [-0.53, +0.24] & \best{43\%} & \best{-0.03} [-0.12, +0.05] & \best{41\%} \\
Rate $2\times10^{-5}$ & \best{+0.13} [-0.29, +0.54] & \best{59\%} & \best{+0.26} [-0.22, +0.74] & \best{59\%} & \best{+0.05} [-0.05, +0.15] & \best{59\%} \\
Rate $5\times10^{-5}$ & \best{+0.66} [-0.55, +1.84] & \best{75\%} & \best{+1.11} [+0.14, +2.05] & \best{77\%} & \best{+0.23} [+0.02, +0.42] & \best{77\%} \\
Rate $2\times10^{-4}$ & \best{+2.11} [-0.22, +4.61] & \best{95\%} & \best{+1.37} [-0.62, +3.27] & \best{80\%} & \best{+0.28} [-0.12, +0.68] & \best{77\%} \\
Rate $3\times10^{-4}$ & \best{+2.09} [-0.75, +4.92] & \best{91\%} & \best{+1.27} [-0.93, +3.65] & \best{74\%} & \best{+0.26} [-0.20, +0.76] & \best{74\%} \\
\bottomrule
\end{tabular}
\end{fitw}
}
\end{table}

\begin{table}[H]
{\revcolor
\caption{\new{Tokens read per article. The opening of an article carries the update.}}
\label{tab:sens-len}
\small
\begin{fitw}{\columnwidth}
\begin{tabular}{l|cc|cc|cc}
\toprule
\multirow{2}{*}[-0.25\normalbaselineskip]{Tokens} & \multicolumn{2}{c|}{IC} & \multicolumn{2}{c|}{Return} & \multicolumn{2}{c}{Sharpe} \\
\cmidrule(lr){2-3}\cmidrule(lr){4-5}\cmidrule(lr){6-7}
 & Gain [95\% CI] & Wins & Gain [95\% CI] & Wins & Gain [95\% CI] & Wins \\
\midrule
64 tokens & \best{+1.94} [+0.38, +3.58] & \best{93\%} & \best{+1.37} [-0.12, +2.83] & \best{82\%} & \best{+0.27} [-0.04, +0.58] & \best{82\%} \\
128 tokens & \best{+1.97} [+0.45, +3.55] & \best{91\%} & \best{+1.50} [-0.00, +2.93] & \best{82\%} & \best{+0.31} [-0.00, +0.60] & \best{77\%} \\
256 tokens & \best{+1.90} [+0.48, +3.37] & \best{89\%} & \best{+1.50} [+0.19, +2.82] & \best{84\%} & \best{+0.31} [+0.03, +0.58] & \best{84\%} \\
512 tokens & \best{+1.80} [+0.28, +3.36] & \best{89\%} & \best{+1.45} [+0.10, +2.76] & \best{86\%} & \best{+0.30} [+0.01, +0.57] & \best{84\%} \\
1024 tokens & \best{+1.80} [+0.32, +3.34] & \best{91\%} & \best{+1.59} [+0.16, +3.01] & \best{82\%} & \best{+0.33} [+0.03, +0.62] & \best{82\%} \\
\bottomrule
\end{tabular}
\end{fitw}
}
\end{table}

\subsection{Continuously Pretrained Suite}
\label{app:pit4b}

\begin{table}[H]
{\revcolor
\caption{\new{PIT-4B on both targets. Snapshots to 2017 are compared with those of 2018 and 2019, which add continued pretraining, and with the snapshots to 2017 carrying an adapter of rank 16 trained for 400 steps at $5\times10^{-5}$ with a batch of 2, the setting of our preliminary runs on this suite. On next-day open-to-close returns every variant falls far below its close-to-close level and the order of the updates changes, showing that \textit{the two targets disagree on this suite}.}}
\label{tab:pit4b}
\small
\setlength{\tabcolsep}{3pt}
\begin{fitw}{\columnwidth}
\begin{tabular}{l|cc|cc}
\toprule
\multirow{2}{*}[-0.25\normalbaselineskip]{Backbone: PIT-4B} & \multicolumn{2}{c|}{Close $\rightarrow$ Next close} & \multicolumn{2}{c}{Next open $\rightarrow$ Next close} \\
\cmidrule(lr){2-3}\cmidrule(lr){4-5}
 & IC & Sharpe & IC & Sharpe \\
\midrule
Snapshots to 2017 (w/o update) & 3.96 & 0.66 & -0.34 & 0.09 \\
Snapshots of 2018 and 2019 (continued pretraining) & 5.49 & 0.89 & 1.08 & -0.01 \\
Snapshots to 2017 + PALM (rank 16) & 6.23 & 0.91 & -0.39 & 0.17 \\
\bottomrule
\end{tabular}
\end{fitw}
}
\end{table}

\new{To check whether the comparison with continued pretraining survives the switch to next-day open-to-close returns, we repeat it on PIT-4B \citep{kelly2026pit}. As shown in Table~\ref{tab:pit4b}, the adapter beats the later snapshots on the close-to-close target, while on next-day open-to-close returns the later snapshots read a higher IC than the adapter, the adapter keeps the higher Sharpe ratio, and every variant falls far below its close-to-close level, so the two targets disagree on this suite. We therefore do not use this suite to compare the two updates.}

\clearpage
\section{\new{Additional Checks}}
\label{app:checks}

\new{\textbf{Same-day return.} To check that the score does not merely repeat the return an article already reports, we read each score against the return realized on its own date, and against the next-day open-to-close return after removing the part explained by the same-day and past-week return. As shown in Table~\ref{tab:chk-sameday}, the score does echo the same-day return, and PALM strengthens that echo, but the gain on the residual target is as large as on the raw target and its interval excludes zero. The gain thus does not come from the move the article describes.}

\begin{table}[H]
{\revcolor
\caption{\new{Same-day return. The IC against the same-day return, the next-day open-to-close return, and that return after removing the same-day and past-week return, showing that \textit{the gain survives the removal of the move the article describes}.}}
\label{tab:chk-sameday}
\small
\setlength{\tabcolsep}{3pt}
\begin{fitw}{\columnwidth}
\begin{tabular}{l|c|c|c}
\toprule
Arm & Same-day return & Next-day return & Residual of next-day return \\
\midrule
Base & 85.11 & 1.63 & 1.59 \\
+ PALM & 95.78 & 3.40 & 3.86 \\
\midrule
Gain [95\% CI] & +10.68 [+7.91, +13.50] & \best{+1.77} [-0.15, +3.71] & \best{+2.27} [+0.44, +4.17] \\
Wins & & \best{93\%} & \best{98\%} \\
\bottomrule
\end{tabular}
\end{fitw}
}
\end{table}

\new{\textbf{Staleness on other windows and recent vintages.} To check that the absence of decay is not specific to the tradable window, we repeat the regression of Table~\ref{tab:stale} on the close-to-close and the overnight returns, and on the vintages at most five years stale. As shown in Table~\ref{tab:chk-stale}, no measure shows a decay we can detect on any window, and the slope among recent vintages leans towards the older ones. Note that the pooled design detects, at the one-sided 5\% level with 80\% power, only a yearly decay larger than 0.35$\times10^{-3}$.}

\begin{table}[H]
{\revcolor
\caption{\new{Staleness on other windows. Slopes are per year of staleness with suite fixed effects and 90\% intervals, showing that \textit{no window and no subset of vintages shows a decay we can detect}.}}
\label{tab:chk-stale}
\small
\setlength{\tabcolsep}{2.5pt}
\begin{fitw}{\columnwidth}
\begin{tabular}{l|c|ccc|ccc|ccc}
\toprule
\multirow{2}{*}[-0.25\normalbaselineskip]{Return window} & \multirow{2}{*}[-0.25\normalbaselineskip]{$n$} & \multicolumn{3}{c|}{IC} & \multicolumn{3}{c|}{Return} & \multicolumn{3}{c}{Sharpe} \\
\cmidrule(lr){3-5}\cmidrule(lr){6-8}\cmidrule(lr){9-11}
 & & Slope & 90\% CI & \begin{tabular}[c]{@{}c@{}}Decay\\detected\end{tabular} & Slope & 90\% CI & \begin{tabular}[c]{@{}c@{}}Decay\\detected\end{tabular} & Slope & 90\% CI & \begin{tabular}[c]{@{}c@{}}Decay\\detected\end{tabular} \\
\midrule
(A) Close $\rightarrow$ Next close & 44 & +0.054 & [-0.185, +0.317] & \xmark & -0.204 & [-0.465, +0.046] & \xmark & -0.033 & [-0.078, +0.010] & \xmark \\
(B) Close $\rightarrow$ Next open & 44 & -0.150 & [-0.395, +0.099] & \xmark & -0.077 & [-0.241, +0.073] & \xmark & -0.021 & [-0.068, +0.023] & \xmark \\
(C) Next open $\rightarrow$ Next close & 44 & +0.088 & [-0.140, +0.321] & \xmark & -0.116 & [-0.296, +0.071] & \xmark & -0.024 & [-0.063, +0.015] & \xmark \\
(C), staleness $\leq$ 5 years & 15 & +0.632 & [-0.302, +1.455] & \xmark & +0.336 & [-0.343, +1.071] & \xmark & +0.070 & [-0.079, +0.227] & \xmark \\
\bottomrule
\end{tabular}
\end{fitw}
}
\end{table}

\new{\textbf{Leave one suite out.} To check that neither finding rests on one suite, we drop each suite in turn. As shown in Table~\ref{tab:chk-loo}, the staleness slope stays without a decay we can detect, and the gain of PALM stays positive on most checkpoints whichever suite is dropped.}

\begin{table}[H]
{\revcolor
\caption{\new{Leave one suite out. The staleness slope of the IC with its 90\% interval and the IC gain of PALM with its 95\% interval, showing that \textit{neither finding rests on a single suite}.}}
\label{tab:chk-loo}
\small
\setlength{\tabcolsep}{3pt}
\begin{fitw}{\columnwidth}
\begin{tabular}{l|c|c|cc}
\toprule
Suites kept & $n$ & Staleness slope [90\% CI] & IC gain [95\% CI] & Wins \\
\midrule
Without ChronoGPT & 23 & +0.242 [-0.032, +0.518] & \best{+1.10} [-0.22, +2.36] & \best{87\%} \\
Without ChronoGPT-Instruct & 28 & +0.038 [-0.223, +0.297] & \best{+2.03} [-0.68, +4.83] & \best{93\%} \\
Without DatedGPT & 37 & +0.073 [-0.156, +0.314] & \best{+1.99} [-0.06, +4.12] & \best{97\%} \\
\bottomrule
\end{tabular}
\end{fitw}
}
\end{table}

\rtwo{\textbf{Further checks.} To check further points on dependence, equivalence and portfolio construction, we compute the quantities in Table~\ref{tab:chk-round2} on the stored scores. The adapted checkpoints show no decay we can detect either, so the update does not trade one stale model for another. The 90\% intervals of the two differences the main text calls undetectable lie within $\pm$1.1 and $\pm$1.2$\times10^{-3}$, which is the margin of equivalence they support, and not a tighter one. Resampling vintages in blocks of three consecutive cutoffs, so that nested continuations are not drawn as independent, leaves the interval of the gain almost unchanged, at a one-sided bootstrap probability of 0.05. A dollar-neutral book, with each leg weighted on its own, keeps the sign of the return and Sharpe gains on most checkpoints with smaller sizes. The gain is somewhat larger on checkpoints with a lower base IC, but stays positive on the upper half. Note that the daily IC itself is weak, with Newey--West $t$-statistics below one, which is why we read the paired gain rather than the level, and why the gain on the IC is our primary measure.}

\begin{table}[H]
{\color{rtwored}
\caption{Further checks. Point estimates with their intervals, in units of $10^{-3}$ for the IC, showing that \textit{the main findings hold under further checks, with the margins stated}.}
\label{tab:chk-round2}
\small
\setlength{\tabcolsep}{3pt}
\begin{fitw}{\columnwidth}
\begin{tabular}{l|c|c|c}
\toprule
Check & Estimate & Interval & Level \\
\midrule
Staleness slope of adapted checkpoints (IC, per year) & +0.078 & [-0.139, +0.283] & 90\% \\
Recency over domain, [B] $-$ [A] (IC) & -0.44 & [-1.11, +0.28] & 90\% \\
PALM $-$ full finetuning (IC) & -0.43 & [-1.22, +0.36] & 90\% \\
\midrule
IC gain, vintages resampled in blocks of three & +1.77 & [-0.29, +3.73] & 95\% \\
Return gain, dollar-neutral book (\%) & +0.73 & [-0.70, +2.14] & 95\% \\
Sharpe gain, dollar-neutral book & +0.16 & [-0.15, +0.46] & 95\% \\
\midrule
Correlation of the IC gain with the base IC & -0.22 & \textemdash & \textemdash \\
Newey--West $t$ of the daily IC, Base / + PALM & 0.48 / 0.99 & \textemdash & \textemdash \\
\bottomrule
\end{tabular}
\end{fitw}
}
\end{table}

\new{\textbf{Leakage audit.} Every adapted panel stores the latest date its adapter read. Across all adapted arms, no panel reads past its decision date, except the 44 panels of the upper bound in Table~\ref{tab:bounds}(b), which do so by design.}

\clearpage
\section{\new{Anticipated Questions}}
\label{app:qa}

{\revcolor
\newcommand{\qa}[2]{\noindent\textbf{Q. #1}\par\noindent A. #2\par\medskip}

\qa{If text from before the cutoff works as well, why call the method point-in-time adaptation?}{The name states the rule the update obeys, not the source of its gain. PALM reads only text published before the decision date, so an adapted checkpoint stays eligible whenever the checkpoint is. Section~\ref{sec:where} then shows that the eligible text after the cutoff is not what drives the gain, which is why the rule has no detectable cost.}

\qa{Is the gain economically meaningful?}{The IC is small in absolute terms, as for most signals read from news, and the daily long-short book does not survive realistic costs for either arm (Table~\ref{tab:cost}). We therefore report the gain as better signal quality, measured by the IC, not as a tradable strategy.}

\qa{The pooled 95\% interval of the gain includes zero. Is the gain real?}{The pooled gain is positive at the one-sided 5\% level, raises the IC on most checkpoints of every suite (Table~\ref{tab:allck}), keeps its sign under every change of protocol (Table~\ref{tab:robust}), and is larger, with an interval above zero, once the same-day and past-week return are removed from the target (Table~\ref{tab:chk-sameday}). Each suite on its own is too small for its interval to exclude zero.}

\qa{Does the score simply repeat the same-day price move the article reports?}{It does echo the same-day return, and PALM strengthens that echo, but the gain on the next-day return after removing the same-day and past-week return is at least as large as on the raw target (Table~\ref{tab:chk-sameday}).}

\qa{Is the absence of decay just a lack of power?}{Partly. The pooled design detects only a yearly decay larger than 0.35$\times10^{-3}$, and the interval rules out one larger than 0.14$\times10^{-3}$ per year. No decay appears on the close-to-close or overnight windows, among vintages at most five years stale, or without any one suite either (Tables~\ref{tab:chk-stale} and~\ref{tab:chk-loo}). A smaller decay, or one confined to the first year, needs suites with more vintages.}

\qa{Why not use retrieval, which raises the IC more?}{Same-firm retrieval changes the input rather than the weights and needs a dated index and a longer prompt for every scored article, and PALM adds little on top of it (Appendix~\ref{app:alt}). The two serve different settings, retrieval when an index can be kept and queried, and PALM when the model itself must carry the domain.}

\qa{How does PALM compare with full finetuning?}{Full finetuning reaches a slightly higher IC, but the paired difference has an interval containing zero on every measure, while PALM trains under 0.5\% of the weights and stores an adapter instead of a full copy of the model for each decision date (Section~\ref{sec:fullft} and Table~\ref{tab:compute}).}

\qa{Why do the selected settings differ so much across backbones?}{Each setting is chosen on a separate validation window, and the pools are small. Note that the gain is positive at every rank of a shared setting (Figure~\ref{fig:capacity}), so the result does not depend on the selected values.}

\qa{Why is DatedGPT weaker?}{Its vintages are trained from scratch on general text and score lowest before any update. Its validation gains are all negative and its gain changes sign across decision dates, so the robustness we claim holds for the pooled gain (Section~\ref{sec:limitations}).}

\qa{Does any adapter read text from the evaluation window?}{No. Every adapted panel stores the latest date its adapter read, and none reads past its decision date, except the 44 panels of the upper bound in Table~\ref{tab:bounds}(b), which do so by design (Appendix~\ref{app:checks}).}

\qa{Does the result carry over to larger models?}{We cannot say. On PIT-4B, the only larger suite, the adapter does not raise the IC on the tradable target (Appendix~\ref{app:pit4b}), and the suites we study are all close to 1.5B parameters in size.}

\qa{Why score next-day open-to-close returns?}{The panel carries a date but no time of day, so an article dated $t$ may appear after the close of $t$. The next-day open-to-close return starts only after every such article is public, and is the first window that is tradable whatever the hour of publication (Section~\ref{sec:align}).}
\rtwo{\qa{Why does the 1999 ChronoGPT vintage score better than the 2019 one?}{A single checkpoint reads a weak signal on a few hundred dates, so its IC is noisy, and the ordering of two vintages says little. The staleness test therefore pools every eligible vintage with a bootstrap over months and checkpoints, and Table~\ref{tab:allck} lists each checkpoint.}}

\rtwo{\qa{The interval rules out a decay larger than 0.14 per year, yet the design detects only one larger than 0.35. Which is it?}{They answer different questions. The first is the largest yearly decay the observed interval excludes, and the second is the decay the design would detect with 80\% power if it were present. Both are in units of $10^{-3}$ (Section~\ref{sec:stale} and Appendix~\ref{app:checks}).}}

\rtwo{\qa{Does staleness matter once the adapter is in place?}{No decay we can detect appears among the adapted checkpoints either (Table~\ref{tab:chk-round2}), so maintaining a suite by updating its checkpoints does not simply replace one stale model with another.}}

\rtwo{\qa{Are "comparable to full finetuning" and "no detectable cost of the eligibility rule" equivalence claims?}{Only at the margin the data support. The 90\% intervals of both differences lie within about $\pm$1.2$\times10^{-3}$ (Table~\ref{tab:chk-round2}), which rules out a large loss but not a small one, so we state them as differences we cannot detect.}}

\rtwo{\qa{Do nested checkpoints inflate the evidence?}{Resampling vintages in blocks of consecutive cutoffs within a suite leaves the interval of the gain almost unchanged (Table~\ref{tab:chk-round2}). The Wins column still counts checkpoints that share the same dates, so we do not read it as a count of independent trials.}}

\rtwo{\qa{Is the book dollar neutral?}{The reported book weights every article by the sign of its demeaned score, so the two legs can differ in size. A dollar-neutral book keeps the sign of the return and Sharpe gains with smaller sizes (Table~\ref{tab:chk-round2}), and we read the IC as the primary measure.}}

}

\clearpage
\section{\new{Full Results}}
\label{app:full}

\new{This section reports, for every result of the main text, the checkpoint as released and each adapted arm on every backbone and pooled over all of them, with the IC, the annualized return and the Sharpe ratio of each arm, and the IC gain and its share of improved checkpoints against the release. The pooled rows are the numbers the main text reports, and Table~\ref{tab:adapt} lists every measure by backbone.}

\newcommand{\fullhead}{%
\toprule
Backbone & Arm & IC & Return & Sharpe & IC gain [95\% CI] & Wins \\
\midrule}

\begin{table}[H]
{\revcolor
\caption{\new{Full results of Table~\ref{tab:compute}. Full finetuning and PALM read the same text at the settings of Appendix~\ref{app:cost}.}}
\label{tab:full-compute}
\small
\setlength{\tabcolsep}{2.5pt}
\begin{fitw}{\columnwidth}
\begin{tabular}{l|l|ccc|cc}
\fullhead
ChronoGPT & Base & 1.67 & 2.01 & 0.42 & & \\
 & + Full finetuning & 4.68 & 4.31 & 0.88 & \best{+3.01} [+0.53, +5.66] & \best{95\%} \\
 & + PALM & 4.17 & 3.65 & 0.76 & \best{+2.50} [-0.77, +5.92] & \best{100\%} \\
\cmidrule(lr){1-7}
ChronoGPT-Instruct & Base & 2.21 & 2.95 & 0.61 & & \\
 & + Full finetuning & 4.02 & 3.50 & 0.70 & \best{+1.81} [-0.39, +3.98] & \best{100\%} \\
 & + PALM & 3.52 & 3.79 & 0.79 & \best{+1.31} [-0.26, +2.99] & \best{94\%} \\
\cmidrule(lr){1-7}
DatedGPT & Base & 0.17 & -0.68 & -0.15 & & \\
 & + Full finetuning & 0.81 & 0.05 & 0.01 & \best{+0.63} [-0.70, +1.92] & \best{71\%} \\
 & + PALM & 0.80 & 0.46 & 0.10 & \best{+0.62} [-1.55, +2.75] & \best{71\%} \\
\cmidrule(lr){1-7}
All backbones & Base & 1.63 & 1.93 & 0.40 & & \\
 & + Full finetuning & 3.83 & 3.34 & 0.67 & \best{+2.20} [+0.36, +4.09] & \best{93\%} \\
 & + PALM & 3.40 & 3.19 & 0.66 & \best{+1.77} [-0.15, +3.71] & \best{93\%} \\
\bottomrule
\end{tabular}
\end{fitw}
}
\end{table}

\begin{table}[H]
{\revcolor
\caption{\new{Full results of Table~\ref{tab:domain} and Figure~\ref{fig:domain}, on the checkpoints both adapters cover.}}
\label{tab:full-domain}
\small
\setlength{\tabcolsep}{2.5pt}
\begin{fitw}{\columnwidth}
\begin{tabular}{l|l|ccc|cc}
\fullhead
ChronoGPT & Base & 1.62 & 3.05 & 0.63 & & \\
 & + [A] Text before the cutoff & 4.50 & 3.55 & 0.70 & \best{+2.88} [-0.44, +6.27] & \best{100\%} \\
 & + [B] Text after the cutoff & 3.23 & 3.30 & 0.65 & \best{+1.61} [-1.66, +4.99] & \best{100\%} \\
\cmidrule(lr){1-7}
ChronoGPT-Instruct & Base & 1.13 & 2.74 & 0.58 & & \\
 & + [A] Text before the cutoff & 3.25 & 4.24 & 0.88 & \best{+2.12} [+0.35, +4.00] & \best{100\%} \\
 & + [B] Text after the cutoff & 3.14 & 4.22 & 0.88 & \best{+2.02} [+0.21, +3.79] & \best{100\%} \\
\cmidrule(lr){1-7}
DatedGPT & Base & -0.14 & -0.85 & -0.18 & & \\
 & + [A] Text before the cutoff & 0.27 & 0.40 & 0.10 & \best{+0.41} [-2.00, +2.87] & \best{83\%} \\
 & + [B] Text after the cutoff & 0.34 & 0.13 & 0.03 & \best{+0.48} [-1.68, +2.89] & \best{67\%} \\
\cmidrule(lr){1-7}
All backbones & Base & 0.87 & 1.65 & 0.34 & & \\
 & + [A] Text before the cutoff & 2.67 & 2.73 & 0.56 & \best{+1.80} [-0.04, +3.56] & \best{94\%} \\
 & + [B] Text after the cutoff & 2.24 & 2.55 & 0.52 & \best{+1.37} [-0.45, +3.10] & \best{89\%} \\
\bottomrule
\end{tabular}
\end{fitw}
}
\end{table}

\begin{table}[H]
{\revcolor
\caption{\new{Full results of Table~\ref{tab:robust}, variants [1] to [3].}}
\label{tab:full-robust}
\small
\setlength{\tabcolsep}{2.5pt}
\begin{fitw}{\columnwidth}
\begin{tabular}{l|l|ccc|cc}
\fullhead
\multicolumn{7}{l}{\textit{[1] Firm-level cross-section}} \\
\midrule
ChronoGPT & Base & 1.41 & 2.61 & 0.56 & & \\
 & + PALM & 3.47 & 3.38 & 0.72 & \best{+2.06} [-1.71, +5.87] & \best{100\%} \\
\cmidrule(lr){1-7}
ChronoGPT-Instruct & Base & 2.23 & 4.11 & 0.87 & & \\
 & + PALM & 3.32 & 4.35 & 0.92 & \best{+1.09} [-0.61, +2.75] & \best{94\%} \\
\cmidrule(lr){1-7}
DatedGPT & Base & 0.42 & -0.63 & -0.14 & & \\
 & + PALM & 1.58 & 0.65 & 0.14 & \best{+1.15} [-1.13, +3.42] & \best{71\%} \\
\cmidrule(lr){1-7}
All backbones & Base & 1.55 & 2.64 & 0.56 & & \\
 & + PALM & 3.11 & 3.30 & 0.70 & \best{+1.56} [-0.69, +3.66] & \best{93\%} \\
\midrule
\midrule
\multicolumn{7}{l}{\textit{[2] Syndicated copies dropped}} \\
\midrule
ChronoGPT & Base & 1.28 & 1.94 & 0.40 & & \\
 & + PALM & 4.36 & 3.91 & 0.80 & \best{+3.09} [-0.38, +6.55] & \best{100\%} \\
\cmidrule(lr){1-7}
ChronoGPT-Instruct & Base & 1.73 & 3.25 & 0.67 & & \\
 & + PALM & 3.10 & 3.96 & 0.81 & \best{+1.37} [-0.22, +3.01] & \best{94\%} \\
\cmidrule(lr){1-7}
DatedGPT & Base & -0.14 & -0.78 & -0.17 & & \\
 & + PALM & 0.82 & 0.51 & 0.11 & \best{+0.96} [-1.26, +3.07] & \best{71\%} \\
\cmidrule(lr){1-7}
All backbones & Base & 1.21 & 1.98 & 0.41 & & \\
 & + PALM & 3.34 & 3.39 & 0.70 & \best{+2.12} [+0.09, +4.14] & \best{93\%} \\
\midrule
\midrule
\multicolumn{7}{l}{\textit{[3] Padding left out of the loss}} \\
\midrule
ChronoGPT & Base & 1.67 & 2.01 & 0.42 & & \\
 & + PALM & 4.05 & 2.92 & 0.58 & \best{+2.38} [-0.94, +5.93] & \best{90\%} \\
\cmidrule(lr){1-7}
ChronoGPT-Instruct & Base & 2.21 & 2.95 & 0.61 & & \\
 & + PALM & 3.38 & 3.45 & 0.69 & \best{+1.17} [-0.65, +3.03] & \best{81\%} \\
\cmidrule(lr){1-7}
DatedGPT & Base & 0.17 & -0.68 & -0.15 & & \\
 & + PALM & 0.96 & 0.04 & 0.00 & \best{+0.78} [-1.32, +2.84] & \best{71\%} \\
\cmidrule(lr){1-7}
All backbones & Base & 1.63 & 1.93 & 0.40 & & \\
 & + PALM & 3.32 & 2.65 & 0.53 & \best{+1.69} [-0.34, +3.80] & \best{84\%} \\
\bottomrule
\end{tabular}
\end{fitw}
}
\end{table}

\begin{table}[H]
{\revcolor
\caption{\new{Full results of Table~\ref{tab:robust}, variant [4], with the adapter refitted with three other seeds.}}
\label{tab:full-seeds}
\small
\setlength{\tabcolsep}{2.5pt}
\begin{fitw}{\columnwidth}
\begin{tabular}{l|l|ccc|cc}
\fullhead
ChronoGPT & Base & 1.67 & 2.01 & 0.42 & & \\
 & + PALM, seed 1 & 3.89 & 3.19 & 0.65 & \best{+2.22} [-0.73, +5.37] & \best{90\%} \\
 & + PALM, seed 2 & 4.41 & 3.42 & 0.70 & \best{+2.74} [-0.48, +6.03] & \best{95\%} \\
 & + PALM, seed 3 & 4.06 & 3.48 & 0.71 & \best{+2.39} [-0.61, +5.59] & \best{100\%} \\
\cmidrule(lr){1-7}
ChronoGPT-Instruct & Base & 2.21 & 2.95 & 0.61 & & \\
 & + PALM, seed 1 & 3.42 & 3.66 & 0.76 & \best{+1.21} [-0.43, +2.87] & \best{94\%} \\
 & + PALM, seed 2 & 3.52 & 3.40 & 0.70 & \best{+1.31} [-0.33, +3.00] & \best{94\%} \\
 & + PALM, seed 3 & 3.40 & 3.78 & 0.78 & \best{+1.19} [-0.41, +2.83] & \best{88\%} \\
\cmidrule(lr){1-7}
DatedGPT & Base & 0.17 & -0.68 & -0.15 & & \\
 & + PALM, seed 1 & 0.51 & 0.42 & 0.10 & \best{+0.34} [-1.82, +2.56] & \best{71\%} \\
 & + PALM, seed 2 & 0.76 & 0.23 & 0.06 & \best{+0.59} [-1.73, +2.93] & \best{86\%} \\
 & + PALM, seed 3 & 0.53 & 0.41 & 0.09 & \best{+0.36} [-1.64, +2.26] & \best{71\%} \\
\cmidrule(lr){1-7}
All backbones & Base & 1.63 & 1.93 & 0.40 & & \\
 & + PALM, seed 1 & 3.18 & 2.92 & 0.60 & \best{+1.55} [-0.26, +3.37] & \best{89\%} \\
 & + PALM, seed 2 & 3.51 & 2.91 & 0.60 & \best{+1.88} [-0.02, +3.85] & \best{93\%} \\
 & + PALM, seed 3 & 3.26 & 3.10 & 0.64 & \best{+1.63} [-0.18, +3.43] & \best{91\%} \\
\bottomrule
\end{tabular}
\end{fitw}
}
\end{table}

\begin{table}[H]
{\revcolor
\caption{\new{Full results of Table~\ref{tab:robust}, variant [5], with the whole experiment repeated at each decision date at the shared setting.}}
\label{tab:full-dates}
\small
\setlength{\tabcolsep}{2.5pt}
\begin{fitw}{\columnwidth}
\begin{tabular}{l|l|ccc|cc}
\fullhead
\multicolumn{7}{l}{\textit{Decision date November 2016}} \\
\midrule
ChronoGPT & Base & 2.67 & -0.21 & -0.04 & & \\
 & + PALM & 3.52 & 0.66 & 0.14 & \best{+0.85} [-0.45, +2.17] & \best{88\%} \\
\cmidrule(lr){1-7}
ChronoGPT-Instruct & Base & 3.78 & -0.44 & -0.09 & & \\
 & + PALM & 5.13 & 0.27 & 0.06 & \best{+1.36} [+0.11, +2.63] & \best{100\%} \\
\cmidrule(lr){1-7}
DatedGPT & Base & 3.84 & 0.06 & 0.02 & & \\
 & + PALM & 3.71 & 0.07 & 0.03 & \best{-0.14} [-1.76, +1.73] & \best{33\%} \\
\cmidrule(lr){1-7}
All backbones & Base & 3.19 & -0.27 & -0.06 & & \\
 & + PALM & 4.14 & 0.46 & 0.10 & \best{+0.95} [-0.07, +2.00] & \best{88\%} \\
\midrule
\midrule
\multicolumn{7}{l}{\textit{Decision date November 2017}} \\
\midrule
ChronoGPT & Base & 1.01 & -1.15 & -0.23 & & \\
 & + PALM & 2.60 & 0.64 & 0.13 & \best{+1.59} [+0.24, +3.05] & \best{100\%} \\
\cmidrule(lr){1-7}
ChronoGPT-Instruct & Base & 2.66 & -0.68 & -0.14 & & \\
 & + PALM & 4.38 & 0.45 & 0.09 & \best{+1.71} [+0.41, +3.14] & \best{100\%} \\
\cmidrule(lr){1-7}
DatedGPT & Base & 3.24 & -0.21 & -0.04 & & \\
 & + PALM & 2.71 & -1.06 & -0.21 & \best{-0.52} [-2.67, +1.62] & \best{50\%} \\
\cmidrule(lr){1-7}
All backbones & Base & 1.88 & -0.87 & -0.18 & & \\
 & + PALM & 3.27 & 0.38 & 0.07 & \best{+1.40} [+0.32, +2.52] & \best{94\%} \\
\midrule
\midrule
\multicolumn{7}{l}{\textit{Decision date November 2018}} \\
\midrule
ChronoGPT & Base & 2.47 & -0.43 & -0.09 & & \\
 & + PALM & 3.83 & 1.15 & 0.22 & \best{+1.35} [-0.15, +2.89] & \best{95\%} \\
\cmidrule(lr){1-7}
ChronoGPT-Instruct & Base & 3.21 & 0.06 & 0.01 & & \\
 & + PALM & 4.46 & 0.52 & 0.10 & \best{+1.26} [-0.20, +2.68] & \best{100\%} \\
\cmidrule(lr){1-7}
DatedGPT & Base & 3.61 & 0.12 & 0.03 & & \\
 & + PALM & 3.25 & -0.72 & -0.14 & \best{-0.35} [-1.96, +1.32] & \best{40\%} \\
\cmidrule(lr){1-7}
All backbones & Base & 2.89 & -0.18 & -0.04 & & \\
 & + PALM & 3.99 & 0.67 & 0.13 & \best{+1.09} [-0.05, +2.19] & \best{89\%} \\
\bottomrule
\end{tabular}
\end{fitw}
}
\end{table}

\begin{table}[H]
{\revcolor
\addtocounter{table}{-1}
\caption{\new{Full results of Table~\ref{tab:robust}, variant [5] (continued).}}
\small
\setlength{\tabcolsep}{2.5pt}
\begin{fitw}{\columnwidth}
\begin{tabular}{l|l|ccc|cc}
\fullhead
\multicolumn{7}{l}{\textit{Decision date November 2019}} \\
\midrule
ChronoGPT & Base & -1.75 & -1.80 & -0.34 & & \\
 & + PALM & 0.39 & 0.49 & 0.09 & \best{+2.14} [+0.42, +3.96] & \best{100\%} \\
\cmidrule(lr){1-7}
ChronoGPT-Instruct & Base & -0.75 & -0.90 & -0.17 & & \\
 & + PALM & 0.87 & -0.26 & -0.05 & \best{+1.61} [+0.14, +3.14] & \best{100\%} \\
\cmidrule(lr){1-7}
DatedGPT & Base & 0.27 & -1.53 & -0.31 & & \\
 & + PALM & -0.42 & -2.14 & -0.43 & \best{-0.69} [-2.35, +0.88] & \best{17\%} \\
\cmidrule(lr){1-7}
All backbones & Base & -1.09 & -1.43 & -0.27 & & \\
 & + PALM & 0.44 & -0.17 & -0.04 & \best{+1.53} [+0.30, +2.75] & \best{88\%} \\
\midrule
\midrule
\multicolumn{7}{l}{\textit{Decision date November 2021}} \\
\midrule
ChronoGPT & Base & -0.57 & -0.15 & -0.04 & & \\
 & + PALM & 1.73 & 1.82 & 0.37 & \best{+2.30} [-0.05, +4.63] & \best{95\%} \\
\cmidrule(lr){1-7}
ChronoGPT-Instruct & Base & -0.16 & 0.46 & 0.09 & & \\
 & + PALM & 0.92 & 1.58 & 0.31 & \best{+1.08} [-0.83, +2.93] & \best{76\%} \\
\cmidrule(lr){1-7}
DatedGPT & Base & -1.59 & -2.35 & -0.52 & & \\
 & + PALM & -0.19 & -0.13 & -0.03 & \best{+1.40} [-1.09, +3.85] & \best{88\%} \\
\cmidrule(lr){1-7}
All backbones & Base & -0.60 & -0.31 & -0.07 & & \\
 & + PALM & 1.11 & 1.41 & 0.28 & \best{+1.71} [+0.05, +3.28] & \best{87\%} \\
\midrule
\midrule
\multicolumn{7}{l}{\textit{Decision date November 2022}} \\
\midrule
ChronoGPT & Base & -0.41 & 0.81 & 0.20 & & \\
 & + PALM & 1.59 & 2.05 & 0.50 & \best{+2.00} [-1.05, +5.06] & \best{91\%} \\
\cmidrule(lr){1-7}
ChronoGPT-Instruct & Base & -0.77 & 0.48 & 0.11 & & \\
 & + PALM & 0.12 & 2.58 & 0.61 & \best{+0.89} [-1.39, +3.15] & \best{76\%} \\
\cmidrule(lr){1-7}
DatedGPT & Base & -0.47 & -0.07 & -0.02 & & \\
 & + PALM & 0.77 & 1.49 & 0.39 & \best{+1.23} [-2.17, +4.59] & \best{50\%} \\
\cmidrule(lr){1-7}
All backbones & Base & -0.55 & 0.54 & 0.13 & & \\
 & + PALM & 0.92 & 2.15 & 0.52 & \best{+1.47} [-0.67, +3.61] & \best{79\%} \\
\bottomrule
\end{tabular}
\end{fitw}
}
\end{table}

\begin{table}[H]
{\revcolor
\caption{\new{Full results of Figure~\ref{fig:capacity}, at the shared setting with the rank varied.}}
\label{tab:full-rank}
\small
\setlength{\tabcolsep}{2.5pt}
\begin{fitw}{\columnwidth}
\begin{tabular}{l|l|ccc|cc}
\fullhead
ChronoGPT & Base & 1.70 & 2.20 & 0.46 & & \\
 & + PALM, rank 1 & 3.28 & 3.76 & 0.79 & \best{+1.57} [-0.34, +3.52] & \best{95\%} \\
 & + PALM, rank 2 & 4.24 & 4.34 & 0.91 & \best{+2.54} [+0.58, +4.54] & \best{95\%} \\
 & + PALM, rank 3 & 4.38 & 4.44 & 0.93 & \best{+2.68} [+0.40, +5.02] & \best{100\%} \\
 & + PALM, rank 4 & 4.31 & 4.33 & 0.90 & \best{+2.61} [+0.08, +5.24] & \best{95\%} \\
 & + PALM, rank 6 & 4.54 & 4.34 & 0.90 & \best{+2.84} [+0.07, +5.77] & \best{100\%} \\
 & + PALM, rank 8 & 4.62 & 4.37 & 0.92 & \best{+2.92} [-0.11, +6.03] & \best{100\%} \\
 & + PALM, rank 12 & 4.33 & 3.80 & 0.79 & \best{+2.63} [-0.50, +5.90] & \best{95\%} \\
 & + PALM, rank 16 & 4.36 & 3.49 & 0.72 & \best{+2.65} [-0.73, +6.09] & \best{95\%} \\
 & + PALM, rank 24 & 4.07 & 3.37 & 0.71 & \best{+2.37} [-0.83, +5.93] & \best{90\%} \\
 & + PALM, rank 32 & 3.75 & 3.06 & 0.63 & \best{+2.05} [-1.16, +5.45] & \best{95\%} \\
\cmidrule(lr){1-7}
ChronoGPT-Instruct & Base & 2.28 & 2.81 & 0.58 & & \\
 & + PALM, rank 1 & 3.45 & 3.78 & 0.78 & \best{+1.16} [-0.46, +2.80] & \best{88\%} \\
 & + PALM, rank 2 & 3.88 & 3.90 & 0.80 & \best{+1.60} [-0.08, +3.36] & \best{94\%} \\
 & + PALM, rank 3 & 3.98 & 3.93 & 0.81 & \best{+1.70} [-0.42, +3.80] & \best{94\%} \\
 & + PALM, rank 4 & 4.24 & 4.01 & 0.82 & \best{+1.96} [-0.38, +4.14] & \best{94\%} \\
 & + PALM, rank 6 & 4.37 & 3.86 & 0.79 & \best{+2.09} [-0.62, +4.61] & \best{100\%} \\
 & + PALM, rank 8 & 4.23 & 3.61 & 0.74 & \best{+1.94} [-0.91, +4.72] & \best{100\%} \\
 & + PALM, rank 12 & 4.11 & 3.42 & 0.69 & \best{+1.83} [-1.39, +4.94] & \best{88\%} \\
 & + PALM, rank 16 & 4.06 & 2.98 & 0.59 & \best{+1.78} [-1.54, +4.98] & \best{94\%} \\
 & + PALM, rank 24 & 3.99 & 2.98 & 0.60 & \best{+1.71} [-1.87, +5.06] & \best{88\%} \\
 & + PALM, rank 32 & 3.76 & 3.13 & 0.62 & \best{+1.48} [-2.05, +4.94] & \best{94\%} \\
\cmidrule(lr){1-7}
DatedGPT & Base & 0.24 & -0.58 & -0.13 & & \\
 & + PALM, rank 1 & 0.66 & 0.16 & 0.05 & \best{+0.42} [-1.04, +1.90] & \best{71\%} \\
 & + PALM, rank 2 & 0.93 & -0.05 & -0.01 & \best{+0.69} [-0.90, +2.39] & \best{57\%} \\
 & + PALM, rank 3 & 1.09 & 0.68 & 0.16 & \best{+0.85} [-1.31, +3.00] & \best{86\%} \\
 & + PALM, rank 4 & 1.20 & 0.87 & 0.20 & \best{+0.96} [-1.15, +3.22] & \best{86\%} \\
 & + PALM, rank 6 & 1.04 & 0.66 & 0.15 & \best{+0.80} [-1.59, +3.18] & \best{86\%} \\
 & + PALM, rank 8 & 0.48 & 0.36 & 0.08 & \best{+0.24} [-1.06, +1.60] & \best{57\%} \\
 & + PALM, rank 12 & 1.79 & 1.58 & 0.35 & \best{+1.55} [-1.15, +4.49] & \best{86\%} \\
 & + PALM, rank 16 & 1.86 & 1.46 & 0.32 & \best{+1.62} [-1.31, +4.79] & \best{86\%} \\
 & + PALM, rank 24 & 1.83 & 1.77 & 0.38 & \best{+1.59} [-1.64, +5.12] & \best{86\%} \\
 & + PALM, rank 32 & 1.85 & 1.76 & 0.39 & \best{+1.61} [-1.85, +5.31] & \best{86\%} \\
\cmidrule(lr){1-7}
All backbones & Base & 1.68 & 1.98 & 0.41 & & \\
 & + PALM, rank 1 & 2.92 & 3.20 & 0.67 & \best{+1.24} [-0.11, +2.61] & \best{89\%} \\
 & + PALM, rank 2 & 3.58 & 3.48 & 0.72 & \best{+1.90} [+0.48, +3.37] & \best{89\%} \\
 & + PALM, rank 3 & 3.71 & 3.65 & 0.76 & \best{+2.03} [+0.27, +3.83] & \best{95\%} \\
 & + PALM, rank 4 & 3.79 & 3.66 & 0.76 & \best{+2.11} [+0.13, +4.19] & \best{93\%} \\
 & + PALM, rank 6 & 3.92 & 3.58 & 0.74 & \best{+2.24} [-0.04, +4.57] & \best{98\%} \\
 & + PALM, rank 8 & 3.82 & 3.46 & 0.72 & \best{+2.14} [-0.20, +4.52] & \best{93\%} \\
 & + PALM, rank 12 & 3.85 & 3.31 & 0.68 & \best{+2.17} [-0.44, +4.88] & \best{91\%} \\
 & + PALM, rank 16 & 3.85 & 2.98 & 0.61 & \best{+2.17} [-0.50, +4.90] & \best{93\%} \\
 & + PALM, rank 24 & 3.68 & 2.97 & 0.61 & \best{+2.00} [-0.72, +4.80] & \best{89\%} \\
 & + PALM, rank 32 & 3.45 & 2.88 & 0.59 & \best{+1.77} [-1.03, +4.64] & \best{93\%} \\
\bottomrule
\end{tabular}
\end{fitw}
}
\end{table}

\begin{table}[H]
{\revcolor
\caption{\new{Full results of Table~\ref{tab:align}, with every measure read on the return window of each block.}}
\label{tab:full-align}
\small
\setlength{\tabcolsep}{2.5pt}
\begin{fitw}{\columnwidth}
\begin{tabular}{l|l|ccc|cc}
\fullhead
\multicolumn{7}{l}{\textit{(A) Close $\rightarrow$ Next close}} \\
\midrule
ChronoGPT & Base & 7.31 & 7.89 & 1.37 & & \\
 & + PALM & 8.59 & 7.76 & 1.33 & \best{+1.28} [-1.70, +4.52] & \best{81\%} \\
\cmidrule(lr){1-7}
ChronoGPT-Instruct & Base & 9.19 & 9.23 & 1.58 & & \\
 & + PALM & 9.87 & 8.81 & 1.48 & \best{+0.68} [-0.90, +2.25] & \best{88\%} \\
\cmidrule(lr){1-7}
DatedGPT & Base & 5.46 & 3.09 & 0.57 & & \\
 & + PALM & 7.04 & 4.50 & 0.82 & \best{+1.58} [-0.83, +3.85] & \best{86\%} \\
\cmidrule(lr){1-7}
All backbones & Base & 7.70 & 7.62 & 1.32 & & \\
 & + PALM & 8.81 & 7.62 & 1.30 & \best{+1.11} [-0.65, +2.99] & \best{84\%} \\
\midrule
\midrule
\multicolumn{7}{l}{\textit{(B) Close $\rightarrow$ Next open}} \\
\midrule
ChronoGPT & Base & 14.73 & 5.12 & 1.49 & & \\
 & + PALM & 14.87 & 3.60 & 1.05 & \best{+0.13} [-3.62, +4.05] & \best{52\%} \\
\cmidrule(lr){1-7}
ChronoGPT-Instruct & Base & 14.93 & 5.03 & 1.46 & & \\
 & + PALM & 14.72 & 3.83 & 1.08 & \best{-0.21} [-1.60, +1.29] & \best{44\%} \\
\cmidrule(lr){1-7}
DatedGPT & Base & 10.88 & 3.27 & 0.97 & & \\
 & + PALM & 12.82 & 3.57 & 1.07 & \best{+1.94} [-0.56, +4.81] & \best{86\%} \\
\cmidrule(lr){1-7}
All backbones & Base & 14.19 & 4.79 & 1.39 & & \\
 & + PALM & 14.49 & 3.68 & 1.06 & \best{+0.29} [-1.84, +2.48] & \best{55\%} \\
\midrule
\midrule
\multicolumn{7}{l}{\textit{(C) Next open $\rightarrow$ Next close}} \\
\midrule
ChronoGPT & Base & 1.67 & 2.01 & 0.42 & & \\
 & + PALM & 4.17 & 3.65 & 0.76 & \best{+2.50} [-0.77, +5.92] & \best{100\%} \\
\cmidrule(lr){1-7}
ChronoGPT-Instruct & Base & 2.21 & 2.95 & 0.61 & & \\
 & + PALM & 3.52 & 3.79 & 0.79 & \best{+1.31} [-0.26, +2.99] & \best{94\%} \\
\cmidrule(lr){1-7}
DatedGPT & Base & 0.17 & -0.68 & -0.15 & & \\
 & + PALM & 0.80 & 0.46 & 0.10 & \best{+0.62} [-1.55, +2.75] & \best{71\%} \\
\cmidrule(lr){1-7}
All backbones & Base & 1.63 & 1.93 & 0.40 & & \\
 & + PALM & 3.40 & 3.19 & 0.66 & \best{+1.77} [-0.15, +3.71] & \best{93\%} \\
\midrule
\midrule
\multicolumn{7}{l}{\textit{(D) Next close $\rightarrow$ Close after}} \\
\midrule
ChronoGPT & Base & 4.78 & 3.88 & 0.70 & & \\
 & + PALM & 5.42 & 4.33 & 0.74 & \best{+0.65} [-3.06, +4.44] & \best{71\%} \\
\cmidrule(lr){1-7}
ChronoGPT-Instruct & Base & 3.68 & 2.53 & 0.47 & & \\
 & + PALM & 2.86 & 2.42 & 0.45 & \best{-0.82} [-2.88, +1.20] & \best{38\%} \\
\cmidrule(lr){1-7}
DatedGPT & Base & -2.72 & -1.45 & -0.28 & & \\
 & + PALM & -1.87 & -1.18 & -0.23 & \best{+0.84} [-1.01, +2.64] & \best{86\%} \\
\cmidrule(lr){1-7}
All backbones & Base & 3.19 & 2.54 & 0.46 & & \\
 & + PALM & 3.33 & 2.76 & 0.48 & \best{+0.15} [-2.24, +2.34] & \best{61\%} \\
\bottomrule
\end{tabular}
\end{fitw}
}
\end{table}

\begin{table}[H]
{\revcolor
\caption{\new{Full results of Table~\ref{tab:bounds}.}}
\label{tab:full-bounds}
\small
\setlength{\tabcolsep}{2.5pt}
\begin{fitw}{\columnwidth}
\begin{tabular}{l|l|ccc|cc}
\fullhead
ChronoGPT & Base & 1.67 & 2.01 & 0.42 & & \\
 & + PALM & 4.17 & 3.65 & 0.76 & \best{+2.50} [-0.77, +5.92] & \best{100\%} \\
 & + (a) Shuffled word order & 2.10 & 0.67 & 0.14 & \best{+0.43} [-4.58, +4.99] & \best{52\%} \\
 & + (b) Text past the date & 4.28 & 3.26 & 0.68 & \best{+2.61} [-0.38, +5.70] & \best{100\%} \\
\cmidrule(lr){1-7}
ChronoGPT-Instruct & Base & 2.21 & 2.95 & 0.61 & & \\
 & + PALM & 3.52 & 3.79 & 0.79 & \best{+1.31} [-0.26, +2.99] & \best{94\%} \\
 & + (a) Shuffled word order & 2.83 & 3.46 & 0.71 & \best{+0.62} [-1.12, +2.29] & \best{75\%} \\
 & + (b) Text past the date & 3.14 & 3.66 & 0.76 & \best{+0.93} [-0.61, +2.44] & \best{75\%} \\
\cmidrule(lr){1-7}
DatedGPT & Base & 0.17 & -0.68 & -0.15 & & \\
 & + PALM & 0.80 & 0.46 & 0.10 & \best{+0.62} [-1.55, +2.75] & \best{71\%} \\
 & + (a) Shuffled word order & -0.45 & -1.35 & -0.30 & \best{-0.62} [-3.29, +1.53] & \best{29\%} \\
 & + (b) Text past the date & 1.46 & 0.54 & 0.12 & \best{+1.28} [-1.39, +3.91] & \best{86\%} \\
\cmidrule(lr){1-7}
All backbones & Base & 1.63 & 1.93 & 0.40 & & \\
 & + PALM & 3.40 & 3.19 & 0.66 & \best{+1.77} [-0.15, +3.71] & \best{93\%} \\
 & + (a) Shuffled word order & 1.96 & 1.36 & 0.28 & \best{+0.33} [-2.33, +2.81] & \best{57\%} \\
 & + (b) Text past the date & 3.42 & 2.97 & 0.62 & \best{+1.79} [-0.03, +3.59] & \best{89\%} \\
\bottomrule
\end{tabular}
\end{fitw}
}
\end{table}

\begin{table}[H]
{\revcolor
\caption{\new{Full results of Table~\ref{tab:cost}. Turnover is two-sided and per day, and the Sharpe ratio is read after each one-way cost.}}
\label{tab:full-costs}
\small
\setlength{\tabcolsep}{2.5pt}
\begin{fitw}{\columnwidth}
\begin{tabular}{l|l|c|cccc|c}
\toprule
\multirow{2}{*}[-0.25\normalbaselineskip]{Backbone} & \multirow{2}{*}[-0.25\normalbaselineskip]{Arm} & \multirow{2}{*}[-0.25\normalbaselineskip]{Turnover} & \multicolumn{4}{c|}{Sharpe after one-way cost} & \multirow{2}{*}[-0.25\normalbaselineskip]{Breakeven (bp)} \\
\cmidrule(lr){4-7}
 & & & 0 bp & 1 bp & 2 bp & 5 bp & \\
\midrule
ChronoGPT & Base & 1.62 & 0.42 & -0.45 & -1.31 & -3.91 & 0.49 \\
 & + PALM & 1.62 & 0.76 & -0.09 & -0.94 & -3.48 & 0.89 \\
\cmidrule(lr){1-8}
ChronoGPT-Instruct & Base & 1.62 & 0.61 & -0.24 & -1.09 & -3.64 & 0.72 \\
 & + PALM & 1.62 & 0.79 & -0.06 & -0.90 & -3.44 & 0.93 \\
\cmidrule(lr){1-8}
DatedGPT & Base & 1.62 & -0.15 & -1.05 & -1.96 & -4.68 & -0.17 \\
 & + PALM & 1.62 & 0.10 & -0.79 & -1.67 & -4.33 & 0.11 \\
\cmidrule(lr){1-8}
All backbones & Base & 1.62 & 0.40 & -0.47 & -1.34 & -3.94 & 0.47 \\
 & + PALM & 1.62 & 0.66 & -0.19 & -1.04 & -3.60 & 0.78 \\
\bottomrule
\end{tabular}
\end{fitw}
}
\end{table}

\new{\rtwo{Tables~\ref{tab:full-compute}, \ref{tab:full-domain}, \ref{tab:full-robust}, \ref{tab:full-seeds}, \ref{tab:full-dates}, \ref{tab:full-rank}, \ref{tab:full-align}, \ref{tab:full-bounds} and \ref{tab:full-costs}} follow the order of the main text. The IC is in units of $10^{-3}$, the return is annualized in percent, and every interval comes from the month-block $\times$ checkpoint bootstrap. Table~\ref{tab:stale} already reports the effect of staleness on every measure for every suite, and Table~\ref{tab:allck} lists every checkpoint with all three measures. Table~\ref{tab:words} comes from preliminary runs whose series are not kept.}

\subsection{Every Eligible Checkpoint}
\label{app:all}

\begin{table}[H]
{\revcolor
\caption{\new{Every eligible checkpoint. Each is read without and with PALM, on next-day open-to-close returns at the selected setting. PALM raises the IC on most checkpoints of every suite, showing that \textit{the average is not carried by a few checkpoints}.}}
\label{tab:allck}
\small
\setlength{\tabcolsep}{3pt}
\begin{fitw}{\columnwidth}
\begin{tabular}{l|c|cc|cc|cc}
\toprule
\multirow{2}{*}[-0.25\normalbaselineskip]{Suite} & \multirow{2}{*}[-0.25\normalbaselineskip]{Cutoff} & \multicolumn{2}{c|}{IC} & \multicolumn{2}{c|}{Return} & \multicolumn{2}{c}{Sharpe} \\
\cmidrule(lr){3-4}\cmidrule(lr){5-6}\cmidrule(lr){7-8}
 & & Base & + PALM & Base & + PALM & Base & + PALM \\
\midrule
ChronoGPT & 1999 & 4.76 & \best{5.73} & 0.45 & \best{2.01} & 0.10 & \best{0.46} \\
 & 2000 & 1.10 & \best{2.89} & 0.00 & \best{2.55} & 0.00 & \best{0.56} \\
 & 2001 & -0.03 & \best{3.59} & -3.18 & \best{1.56} & -0.71 & \best{0.35} \\
 & 2002 & 1.77 & \best{6.35} & 0.87 & \best{4.99} & 0.19 & \best{1.11} \\
 & 2003 & -0.69 & \best{3.41} & -0.89 & \best{4.76} & -0.20 & \best{1.04} \\
 & 2004 & 2.91 & \best{3.58} & 4.22 & \best{6.18} & 0.94 & \best{1.34} \\
 & 2005 & 0.21 & \best{3.38} & 1.48 & \best{3.97} & 0.31 & \best{0.85} \\
 & 2006 & 2.51 & \best{5.41} & 2.20 & \best{3.71} & 0.47 & \best{0.75} \\
 & 2007 & 3.21 & \best{6.64} & 4.07 & \best{4.90} & 0.83 & \best{1.04} \\
 & 2008 & 2.04 & \best{4.87} & 1.26 & \best{4.21} & 0.27 & \best{0.86} \\
 & 2009 & 1.79 & \best{4.42} & 1.19 & \best{3.40} & 0.26 & \best{0.67} \\
 & 2010 & 1.07 & \best{3.38} & 2.41 & \best{3.06} & 0.52 & \best{0.62} \\
 & 2011 & 1.09 & \best{4.24} & 2.75 & \best{2.86} & 0.56 & \best{0.59} \\
 & 2012 & 1.58 & \best{5.64} & 3.16 & \best{5.34} & 0.65 & \best{1.06} \\
 & 2013 & 2.09 & \best{4.74} & \best{3.99} & 3.40 & \best{0.82} & 0.68 \\
 & 2014 & 1.60 & \best{4.10} & \best{4.67} & 4.15 & \best{0.95} & 0.82 \\
 & 2015 & 2.01 & \best{2.86} & \best{2.98} & 2.54 & \best{0.61} & 0.49 \\
 & 2016 & 2.25 & \best{3.60} & 2.03 & \best{3.49} & 0.42 & \best{0.69} \\
 & 2017 & 1.63 & \best{2.23} & 2.84 & \best{3.72} & 0.58 & \best{0.71} \\
 & 2018 & 1.53 & \best{2.46} & \best{2.31} & 2.18 & \best{0.49} & 0.44 \\
 & 2019 & 0.68 & \best{4.10} & 3.48 & \best{3.72} & 0.75 & \best{0.78} \\
\midrule
ChronoGPT-Instruct & 2004 & 2.35 & \best{2.76} & 1.48 & \best{2.62} & 0.31 & \best{0.55} \\
 & 2005 & 2.13 & \best{3.33} & \best{2.62} & 2.07 & \best{0.54} & 0.41 \\
 & 2006 & 5.12 & \best{5.97} & \best{3.96} & 2.09 & \best{0.81} & 0.43 \\
 & 2007 & \best{3.78} & 3.58 & \best{4.13} & 3.30 & \best{0.86} & 0.69 \\
 & 2008 & 3.55 & \best{4.22} & 3.03 & \best{3.31} & 0.62 & \best{0.68} \\
 & 2009 & 3.41 & \best{4.54} & 2.42 & \best{3.32} & 0.50 & \best{0.69} \\
 & 2010 & 2.08 & \best{2.98} & 2.10 & \best{4.75} & 0.44 & \best{0.98} \\
 & 2011 & 1.53 & \best{3.69} & 2.27 & \best{5.21} & 0.49 & \best{1.06} \\
 & 2012 & 2.52 & \best{3.01} & \best{4.27} & 3.64 & \best{0.90} & 0.77 \\
 & 2013 & 2.15 & \best{3.43} & 4.53 & \best{4.97} & 0.91 & \best{1.06} \\
 & 2014 & 1.56 & \best{3.39} & 3.31 & \best{4.53} & 0.71 & \best{0.97} \\
 & 2015 & 1.99 & \best{4.00} & 4.29 & \best{4.47} & 0.89 & \best{0.92} \\
 & 2016 & 2.08 & \best{5.02} & 3.39 & \best{4.82} & 0.71 & \best{1.01} \\
 & 2017 & 0.45 & \best{1.68} & 0.42 & \best{2.50} & 0.09 & \best{0.51} \\
 & 2018 & 0.95 & \best{2.01} & 2.35 & \best{4.72} & 0.49 & \best{0.98} \\
 & 2019 & -0.28 & \best{2.77} & 2.70 & \best{4.33} & 0.57 & \best{0.89} \\
\midrule
DatedGPT & 2013 & 2.06 & \best{3.54} & 0.31 & \best{2.43} & 0.07 & \best{0.52} \\
 & 2014 & \best{0.63} & -1.76 & -1.66 & \best{-0.39} & -0.35 & \best{-0.08} \\
 & 2015 & \best{5.20} & 5.00 & 2.81 & \best{3.82} & 0.65 & \best{0.85} \\
 & 2016 & -2.37 & \best{-0.67} & -2.04 & \best{0.62} & -0.47 & \best{0.14} \\
 & 2017 & -3.60 & \best{-3.44} & -1.69 & \best{-1.55} & -0.39 & \best{-0.34} \\
 & 2018 & 0.51 & \best{1.88} & -1.50 & \best{-1.11} & -0.32 & \best{-0.25} \\
 & 2019 & -1.21 & \best{1.02} & -1.01 & \best{-0.63} & -0.22 & \best{-0.13} \\
\midrule
\midrule
\multicolumn{2}{l|}{Wins, ChronoGPT} & \multicolumn{2}{c|}{21/21 (100\%)} & \multicolumn{2}{c|}{17/21 (81\%)} & \multicolumn{2}{c}{17/21 (81\%)} \\
\multicolumn{2}{l|}{Wins, ChronoGPT-Instruct} & \multicolumn{2}{c|}{15/16 (94\%)} & \multicolumn{2}{c|}{12/16 (75\%)} & \multicolumn{2}{c}{12/16 (75\%)} \\
\multicolumn{2}{l|}{Wins, DatedGPT} & \multicolumn{2}{c|}{5/7 (71\%)} & \multicolumn{2}{c|}{7/7 (100\%)} & \multicolumn{2}{c}{7/7 (100\%)} \\
\multicolumn{2}{l|}{Wins, all suites} & \multicolumn{2}{c|}{41/44 (93\%)} & \multicolumn{2}{c|}{36/44 (82\%)} & \multicolumn{2}{c}{36/44 (82\%)} \\
\bottomrule
\end{tabular}
\end{fitw}
}
\end{table}

\new{Table~\ref{tab:allck} lists the paired result for each of the 44 checkpoints that Table~\ref{tab:adapt} averages over, which shows the spread behind the averages. The cutoff of a vintage is the year through which it was trained.}

\end{document}